\documentclass[runningheads]{llncs}
\usepackage{graphicx}
\usepackage{booktabs}
\usepackage[above,below]{placeins}
\usepackage[hyphens]{url}
\usepackage{hyperref}
\usepackage{soul,todonotes}
\usepackage{csquotes}
\usepackage[inline]{enumitem}
\usepackage{versions}\excludeversion{ignore}\excludeversion{exclude}
\begin{document}
\title{Segmenting Dead Sea Scroll Fragments\\ for a Scientific Image Set}
\titlerunning{Segmenting Dead Sea Scroll Fragments}

%\thanks{Supported in part by the DFG Project Qumran Digital: Text and Lexicon, DFG project number 465277421 \url{https://gepris.dfg.de/gepris/projekt/4652774215}.}

\author{Bronson Brown-deVost \inst{1} \and Berat Kurar-Barakat \inst{2} \and Nachum Dershowitz \inst{2}}
%\author{Bronson Brown-deVost\inst{1}\orcidID{0000-0003-3655-7807} \and
%Second Author\inst{2,3}\orcidID{1111-2222-3333-4444} \and
%Third Author\inst{3}\orcidID{2222--3333-4444-5555}}
%
%\authorrunning{Anonymous}

\institute{Göttingen Academy of Sciences and Humanities \and Tel Aviv University}

%\institute{Die Niedersächsische Akademie der Wissenschaften zu Göttingen, Lagarde-Haus, Friedländer Weg 11
%37085 Göttingen, Germany \and
%Springer Heidelberg, Tiergartenstr. 17, 69121 Heidelberg, Germany
\maketitle              % typeset the header of the contribution
\begin{abstract}
%The abstract should briefly summarize the contents of the paper in 15--250 words.
This paper presents a customized pipeline for segmenting manuscript fragments from images curated by the Israel Antiquities Authority (IAA). The images present challenges for standard segmentation methods due to the presence of the ruler, color, and plate number bars, as well as a black background that resembles the ink and varying backing substrates. The proposed pipeline, consisting of four steps, addresses these challenges by isolating and solving each difficulty using custom-tailored methods. Further, the usage of a multi-step pipeline will surely be helpful from a conceptual standpoint for other image segmentation projects that encounter problems that have proven intractable when applying any of the more commonly used segmentation techniques.

In addition, we create a dataset with bar detection and fragment segmentation ground truth and evaluate the pipeline steps qualitatively and quantitatively on it. This dataset is publicly available to support the development of the field. It aims to address the lack of standard sets of fragment images and evaluation metrics and enable researchers to evaluate their methods in a reliable and reproducible manner.

\end{abstract}
\section{Introduction}

In 2011, the Israel Antiquities Authority (IAA) initiated a large-scale digital imaging project to create a digital archive of their collection of Dead Sea Scrolls---the largest collection of Dead Sea Scrolls in the world \cite{LL}.\footnote{There are several other institutions that have digitized scrolls from their own collections, especially the Jordan Museum in Amman \url{https://universes.art/en/art-destinations/jordan/amman/museums/jordan-museum/dead-sea-scrolls}, the Israel Museum in Jerusalem
\url{https://www.imj.org.il/en/wings/shrine-book/dead-sea-scrolls}, the Bibliothèque nationale de France \url{https://gallica.bnf.fr/ark:/12148/btv1b8551261c}, and a few other collections.
These are either less fragmentary than the manuscripts we deal with from the IAA or do not supply high enough quality images.} At the time of this writing, they have  digitized nearly every fragment in their archive containing text. The scrolls,\footnote{We refer to the manuscripts in the IAA collection as ``Dead Sea Scrolls'' throughout this paper, as this is probably the most well-known designation of these ancient remains.
The interested researcher will note that the collection is largely comprised of Hebrew, Aramaic, and a few Greek scrolls from caves in and around the area of Khirbet Qumran on the western shore of the Dead Sea dating to the last few centuries before the common era and briefly into the common era, but the Dead Sea Scrolls proper include a much broader corpus including items that are not scrolls, such as tefillin and mezuzot among other items, and also works from a much broader time period, up into the Islamic era, written in Latin, Nabatean, and Arabic from areas all around the Judean Desert.} which are for the most part fragmentary, were imaged at 12 wavelength data points using a custom multispectral camera rig, built specifically for this purpose. The IAA began publishing these images online shortly after the project began, making the collection publicly available to both scholars and the general public.%
\footnote{\url{https://www.deadseascrolls.org.il}.}

The IAA has recently been reorganizing its database of images so as to make higher resolution versions of the images available via the standard IIIF Image API. In addition, they have an experimental front-end for scholars, which incorporates manual transcriptions, a canvas for arranging fragments, and a facility for manually associating regions in  images with letters in transcriptions.%
\footnote{\url{https://sqe.deadseascrolls.org.il}; see also \cite{10.1628/186870316X14805961757430} and \cite{SQE_THB}.} The high-resolution images---as is---are satisfactory for most research purposes, but it is necessary to work with fragments segmented out of the images in order to test small joins and larger reconstructions. The segmentation approach provided here seeks to address that need.

\section{IAA Image Dataset}

We have access to about 180,000 IAA images in total, which present the recto and verso of nearly 20,000 manuscript fragments in full-color (a composite of the multispectral images), in infrared (924nm), and in infrared with a raking light from the right, and with a raking light from the left---all have a resolution of $\sim$1215 PPI (we did not utilize the full multispectral data, which has not yet been made publicly available). These images are consistent in the camera and lighting placement and always include an XRite color target, a Golden Thread object-level target, and a tag with the fragment's plate number.
These three additional objects in each image may appear in various positions, but do not change their placement within a recto or within a verso series of images for each single fragment, as neither the camera nor the fragment in its layout was moved between the acquisition of images (see Fig.~\ref{fig:bar_detection_result} for an example).

\section{Motivation}

Segmentation of the fragments in the IAA image database is desirable for several reasons. First and foremost, it helps us to better account for the comprehensiveness of our access to the fragments. It enables us to more easily compare the fragments as pictured in the current image database with earlier series of images, mostly from the middle of the last century. This can help reveal fragments that may have been passed over in earlier research or prove useful for detecting earlier images of fragments that provide a view of the fragment at a time when it was in a better preserved state. Secondly, it makes the fragments more directly accessible to further computational analysis, such as letter recognition. Finally, it enables scholars to more easily test hypotheses for joins or combinations by means of digital tooling, rather than resorting to physical manipulation of the fragile remains, which is detrimental to their integrity.

\section{Related Work}

There are a number of large collections of fragmentary manuscripts, besides the Dead Sea Scrolls.
These include the Elephantine papyri,%
\footnote{\url{https://www.smb.museum/en/museums-institutions/aegyptisches-museum-und-papyrussammlung/collection-research/research/erc-project-elephantine-localizing-4000-years-of-cultural-history-texts-and-scripts-from-elephantine-island-in-egypt}.}
the Oxyrhynchus papyri,%
\footnote{\url{http://www.papyrology.ox.ac.uk/POxy}.}
the Cairo Genizah,%
\footnote{\url{https://fgp.genizah.org}; \url{https://geniza.princeton.edu};
\url{https://cudl.lib.cam.ac.uk/collections/genizah/1}.}
 and fragments found in medieval book bindings.%
 \footnote{\url{https://fragmentarium.ms}; \url{https://bwb.hypotheses.org}.}
Most of these would also benefit from the image treatments described here.

For much of the digitization of the Cairo Geniza fragments, a contrasting blue background (akin to the use of a blue/green screen for chroma key compositing in films) was used when capturing images, making segmentation trivial \cite{Geniza}.
Artifacts were also removed from those images~\cite{join}.
Early attempts at segmenting fragment from background for the Dead Sea Scroll images were reported in \cite{8354133,TK}, but were not accurate enough for handling the whole corpus.
The use of convolutional networks for semantic segmentation has become quite standard \cite{krizhevsky2012imagenet,Fully}.

Frequently, fragment images are manually segmented, often using a tool included in image software such as GrabCut \cite{GrabCut};
see \cite{Removing}.
Segmentation is a prerequisite before trying to piece fragments together
(e.g.\@ \cite{Material} for parchement and \cite{papyrus,Papy} for papyrus%
\footnote{Also \url{https://ec.europa.eu/research-and-innovation/en/projects/success-stories/all/solving-papyrus-puzzle}.})
or to register verso and recto images.

Recto and verso manuscript images are aligned by affine transformation in \cite{5306023}.

\section{Problematica}

The process of segmenting the IAA's images has been complicated by
several factors:

\begin{enumerate}
\def\labelenumi{\arabic{enumi}.}
\item
  Two calibration bars and a label are present in every image, but
  generally never at the same location in the image frame from one fragment image set to another. This can cause problems for
  certain segmentation processes when the fragment is either very tiny and hard to detect or when the fragment has a large support material that may touch or overlap with one or another of these markers.
\item
  The black felt background---chosen for its nonreflective property---appears very similar to the ink on the manuscripts, problematizing differentiation between written remains
  and holes in the parchment or papyrus.
\item
  Many fragments are affixed to a modern support medium.
\begin{enumerate*}[label=(\alph*)]
\item
    This may have been done with Japanese tissue paper for conservation purposes.
\item
   Or it may have been done for historical reasons, such as a
    researcher placing multiple individual fragments on a single support either to display a proposed join or to keep fragments in near proximity that they wished to keep together.
\end{enumerate*}
\end{enumerate}

\section{Methodological Overview}

In consideration of these factors that complicate the segmentation operation, we have opted for a customized pipeline approach to the problem of segmenting our corpus of images. Specifically, we did not rely on deep learning methods except for the bar detection task, as bars can be labeled using bounding boxes that are relatively easy to generate. In contrast, aligning, binarization, and segmentation tasks require pixel-level annotation, which can be particularly challenging on high-resolution images. We have organized the steps in this pipeline in such a way as to provide the most generally reliable process for the corpus as a whole and have been working to include some amount of logging and automated evaluation of possibly or probably problematic instances of segmentation. The basic process can be outlined as follows:

\begin{enumerate}
\def\labelenumi{\arabic{enumi}.}
\item
  Isolate the calibration bars and label in the recto and verso images.
\item
  Align the recto image with the mirrored verso image, ignoring the
  calibration bars and label.
\item
  Threshold the infrared recto and the (mirrored and aligned) verso.
\item
  Union those two masks for a maximal fragment mask.
\item
  Detect the backing substrate on recto and (mirrored and aligned) verso.

  \begin{enumerate}
  \def\labelenumii{\alph{enumii}.}
  \item
    Perform an intersection on those two masks to isolate the maximal shared mask of the backing substrate.
  \item
    Perform morphological closing on the mask, because the backing substrate often has a woven texture.
  \end{enumerate}
\item
  Subtract the mask in step 5 from the mask in step 4.
\item
  Get contours from the mask in step 6.
\item
  Keep only contours that overlap with masked regions in both recto/verso masks from step 4.
\item
  Filter out any very tiny segmented objects based on the image resolution. (There may be flecks of parchment or papyrus that are not suitable for further research.)
\end{enumerate}

\section{Detailed Process}

\subsection{New Evaluation Dataset}
The dataset is designed to support the development of fragment segmentation field by providing a standard set of images and evaluation metrics. The dataset contains a total of $139$ fragments, where $100$ fragments are included in the train set, $20$ fragments in the validation set and $19$ fragments in the test set. To ensure the validity of our analysis, we selected the training and testing fragments at random and verified that they were disjoint at the manuscript level. Each set contains both color and infrared images of recto and verso sides of the fragments. The color recto and verso images are annotated with bounding boxes around the bars and labels using the online AI assisted annotation platform Hasty.%
\footnote{\url{http://hasty.ai}.}

Additionally, the test set includes ground truth for the segmentation of recto fragments, stored as one or multiple WKT files for each fragment. Each WKT file contains the coordinates of the boundary contours and the coordinates of any holes inside the boundary polygon for each piece of the fragment in the image. This ground truth information will be useful for evaluating and comparing the performance of different fragment segmentation methods. The dataset is publicly available and is expected to contribute to the advancement of the field by providing a standard benchmark.%
\footnote{\url{http://to-be-made-available}.}

\subsection{Isolate Calibration Bars and Labels}\label{sec:ruler}

IAA images typically contain three distinct types of bars: an Xrite color calibration bar, a Golden Thread object level target bar, and a plate-number label.
These objects are present in both the recto and verso images of a fragment, and their positions are identical in both infrared and color images.
However, in recto and verso images, they appear in different positions.
Since these objects can occur at any location and orientation within the image, isolating them at the beginning of the segmentation process prevents them from contaminating any of the subsequent steps in the pipeline.

\begin{figure}[t]
  \centering
\includegraphics[draft=false,width=0.8\linewidth]{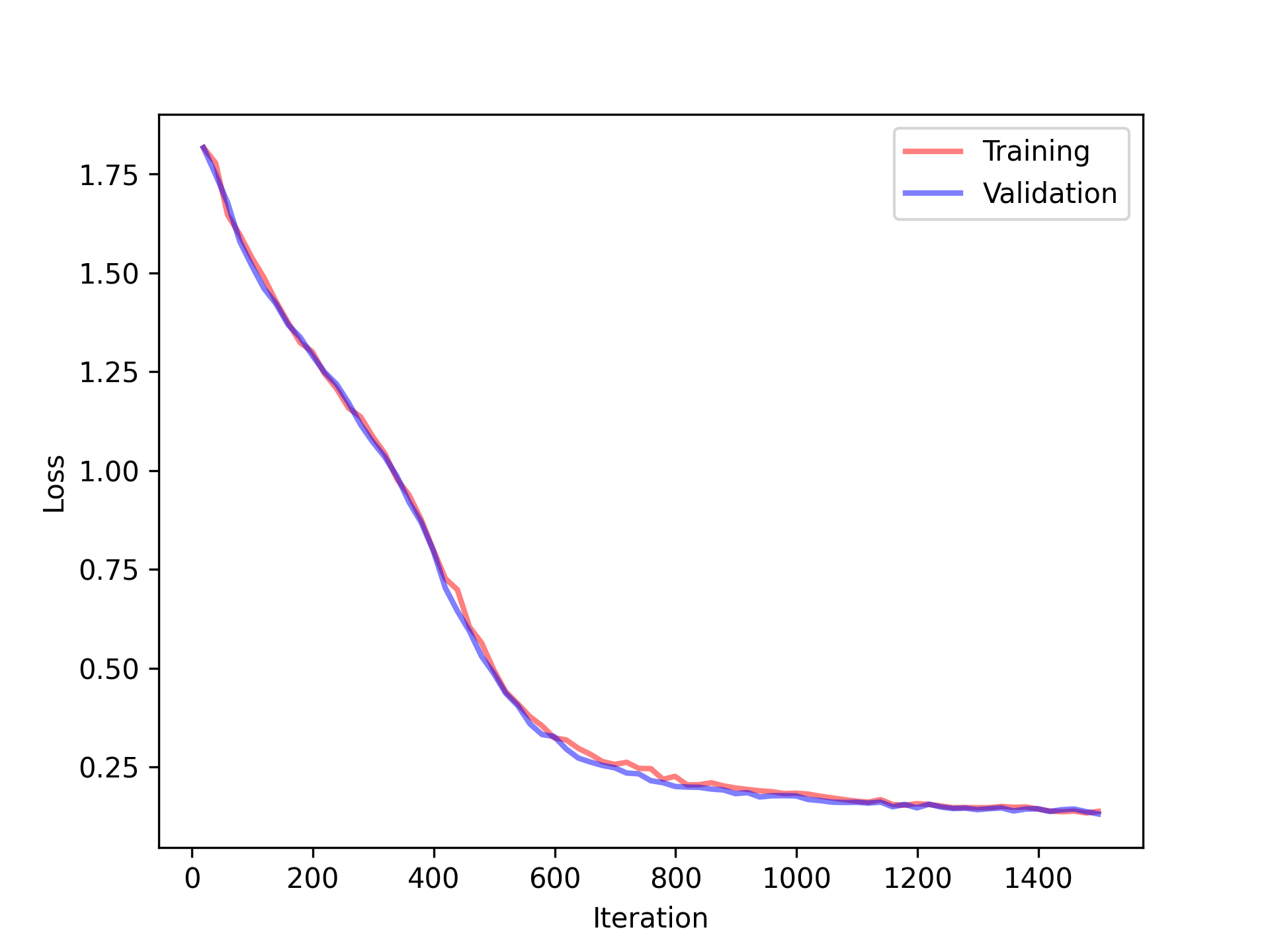} %[scale=1.7]
  \caption{Plot of average classification and regression losses of Faster R-CNN for bar detection on IAA images of fragments versus training and validation iterations.}
  \label{fig:loss}
\end{figure}

To remove the calibration bars and labels, we use a Faster Region-based Convolutional Neural Network (Faster R-CNN) pre-trained on ImageNet but fine-tuned with our own small training set. The motivation behind using the Faster R-CNN method is that it is currently the state-of-the-art detection method in computer vision. Additionally, by using Faster R-CNN, we can avoid the time-consuming task of semantic segmentation or instance segmentation, which require significant effort and time to label accurately, particularly on high-resolution images. The dataset includes both recto and verso color images for each fragment.
We chose to use color images as they contain more separable features for the network.
All types of bars are boxed and labeled as ``bar'' because it is not necessary for our current purposes to identify the bar type.
The region of interest (RoI) heads had a batch size per image of $128$ and were set to only recognize one class (bar).
The images are rescaled so that their shorter side is $800$ pixels.
The network was set to have a base learning rate of $0.0001$ and was trained for $1500$ iterations on the $100$ images training set without decaying the learning rate, with five images per batch, and evaluated on the $20$ images evaluation set with a maximum of five detections per image at every ten iterations observing that the network is not overfitting (Fig.~\ref{fig:loss}).

\begin{figure}[t]
\centering
\begin{tabular}{c c c}
& \textbf{Color} & \textbf{Infrared} \\
\rotatebox{90}{\textbf{Recto}} & \includegraphics[draft=false,width=0.48\textwidth]{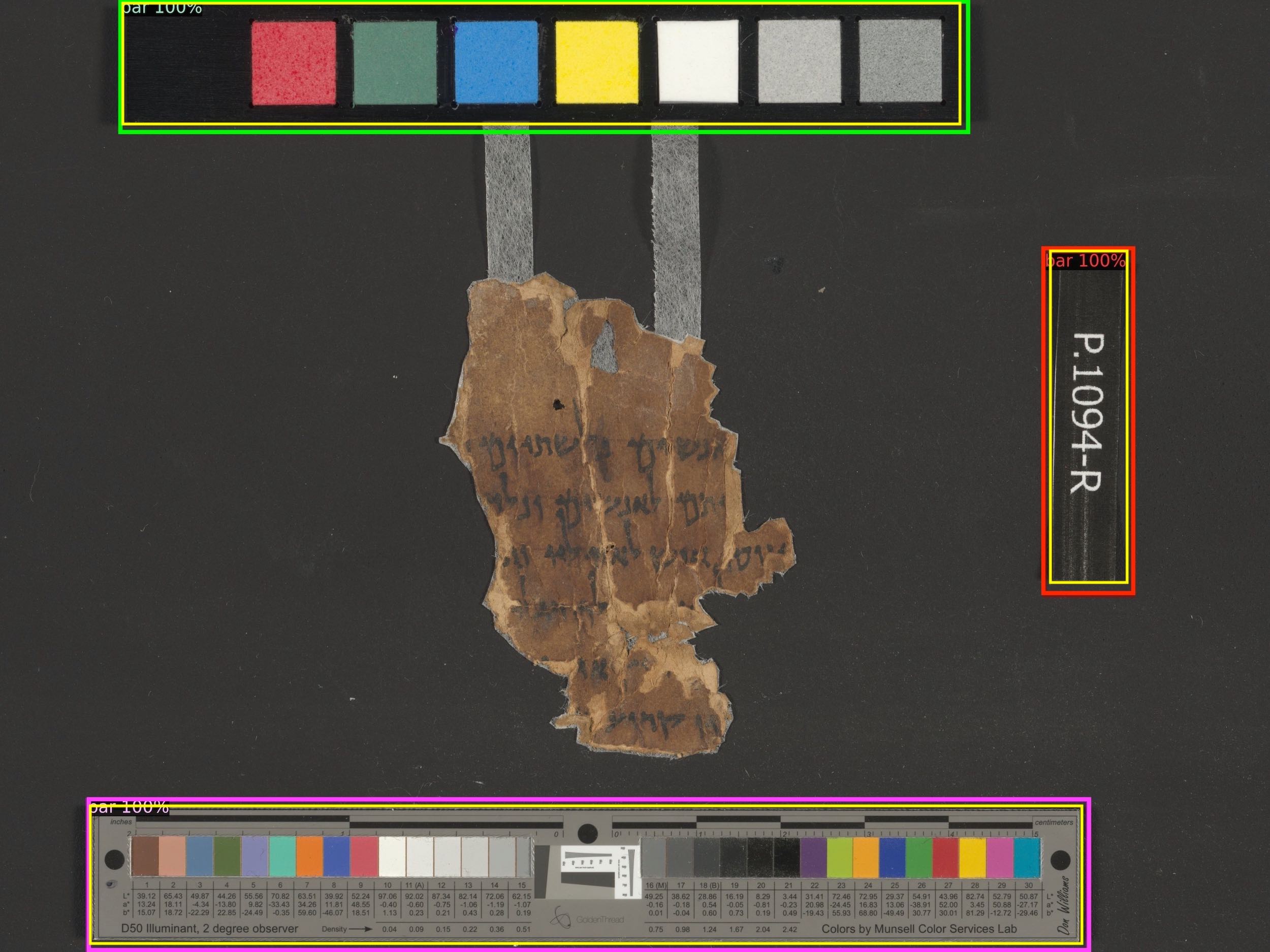} & \includegraphics[draft=false,width=0.48\textwidth]{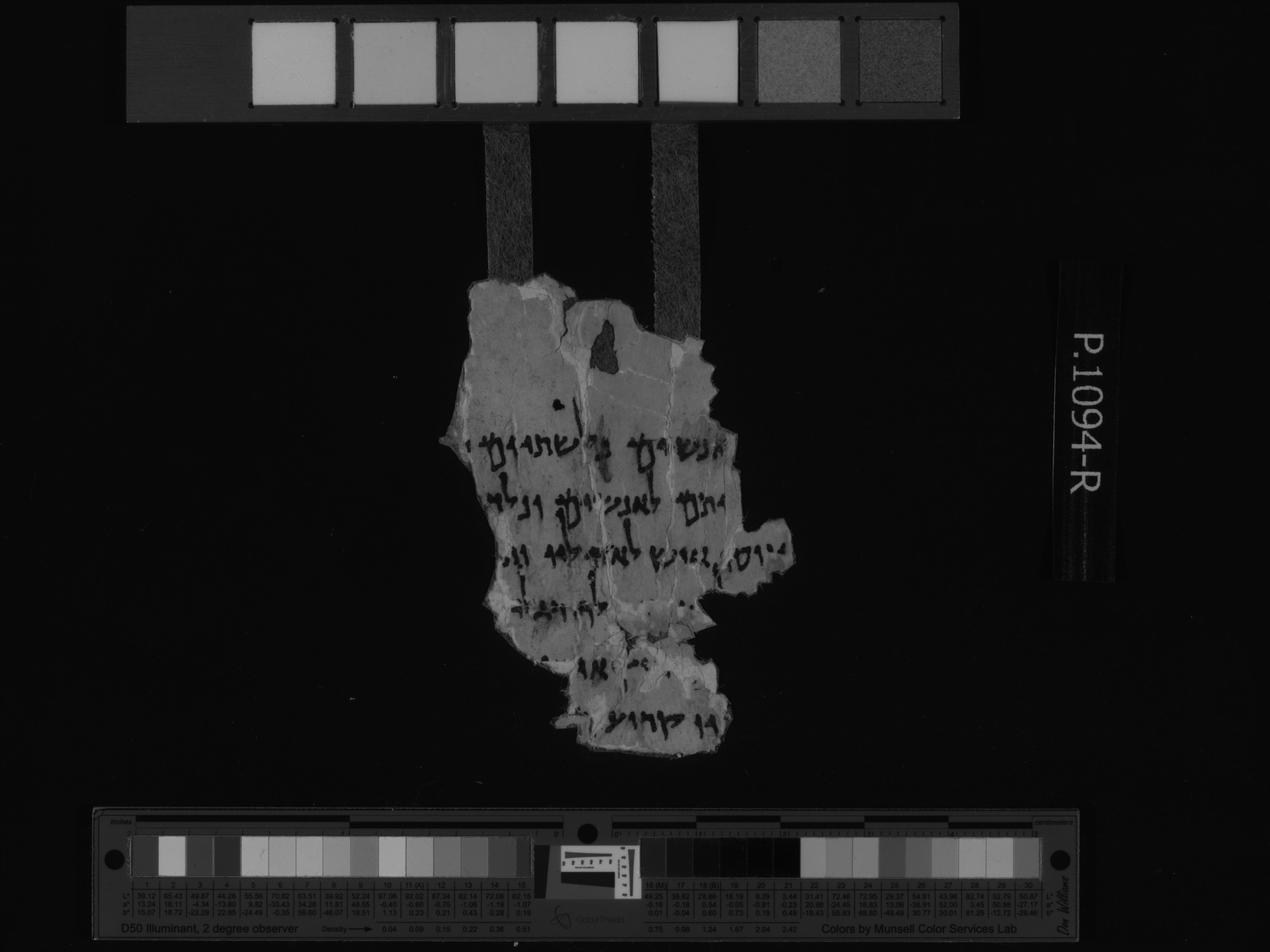} \\
\rotatebox{90}{\textbf{Verso}} & \includegraphics[draft=false,width=0.48\textwidth]{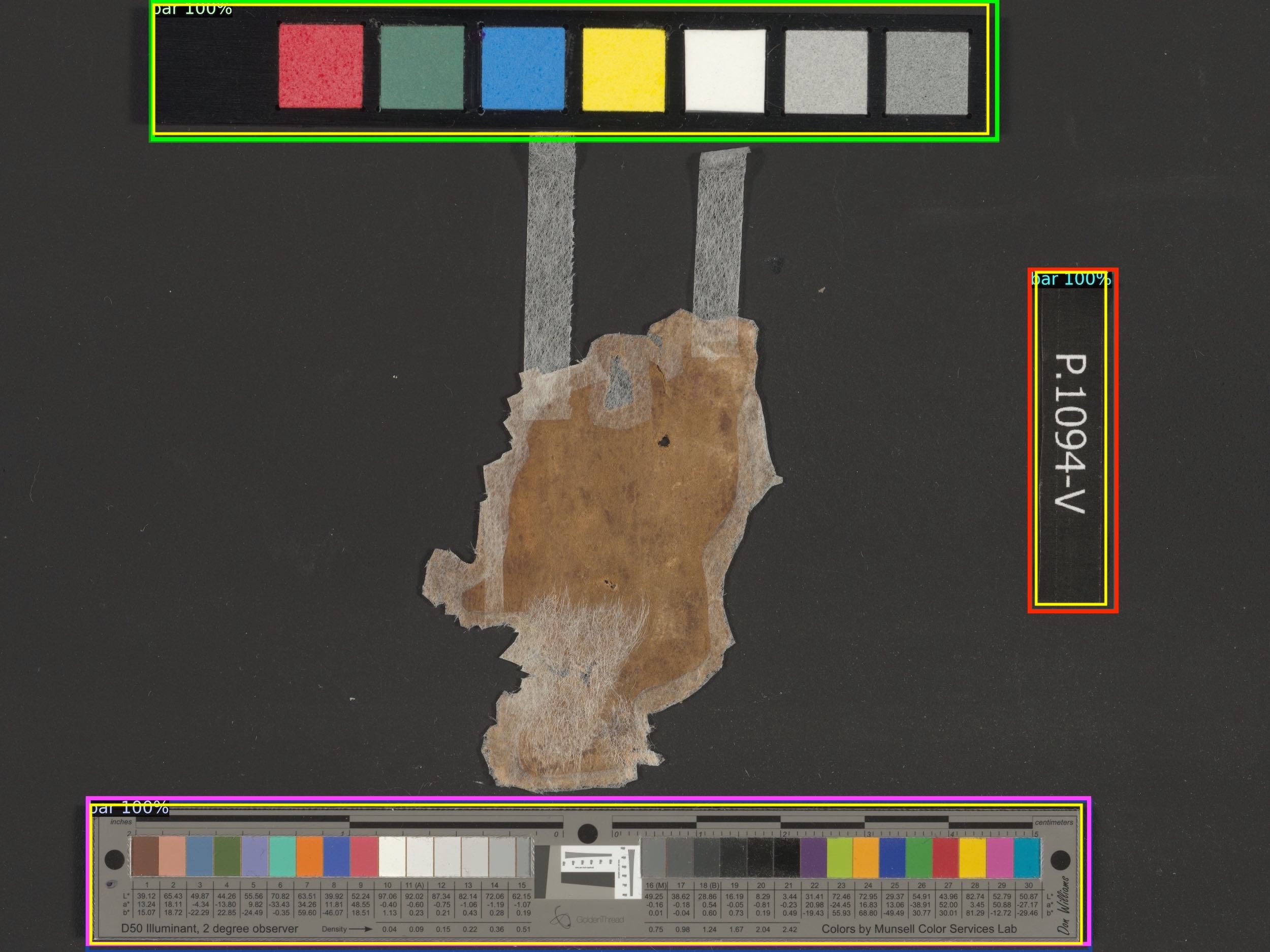} & \includegraphics[draft=false,width=0.48\textwidth]{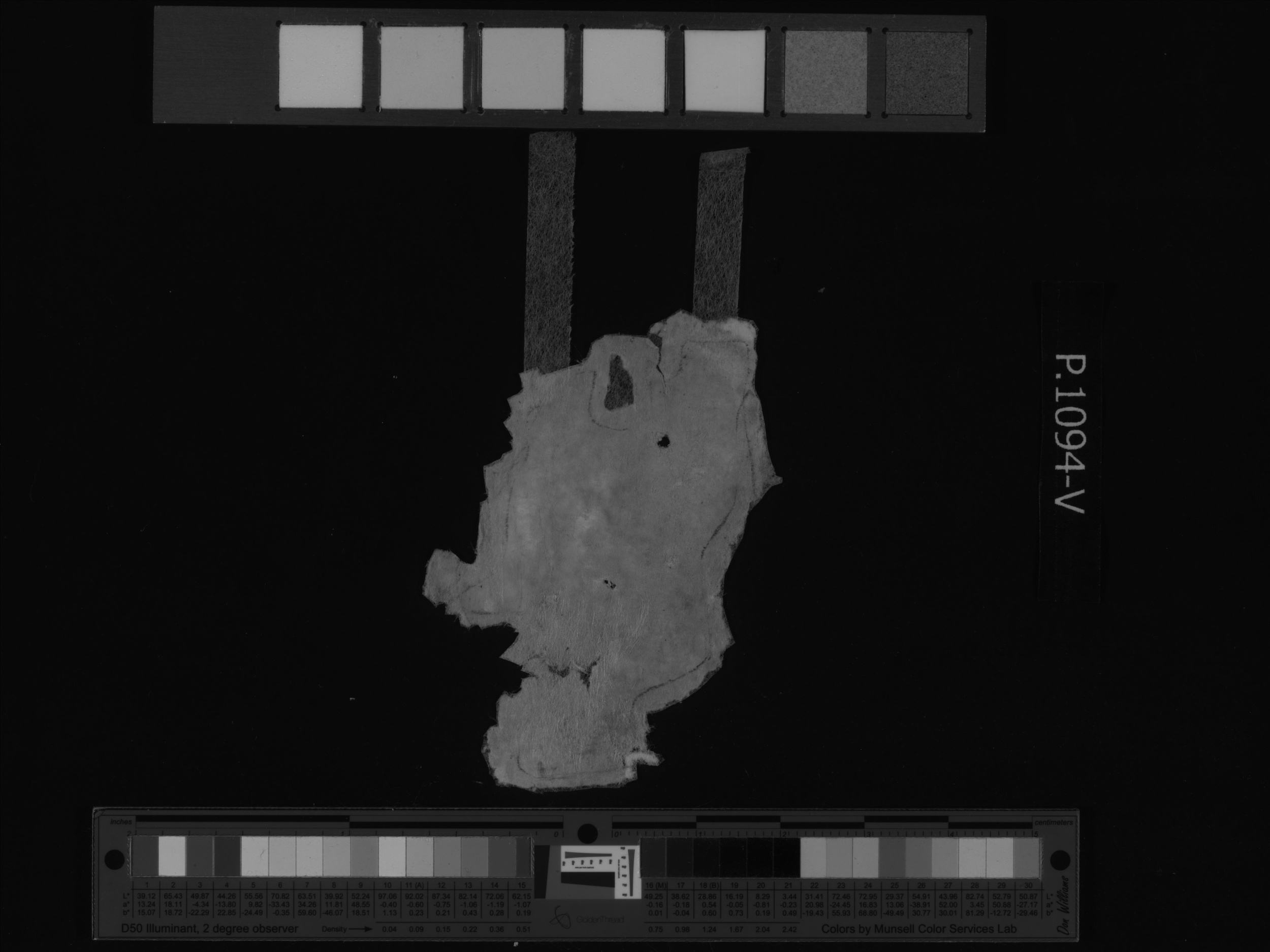} \\
\end{tabular}
\caption{Color images of both recto and verso of a fragment on the  left, with bar detection results overlaid. The images are also overlaid with ground truth in yellow.
The bar detection was performed only on color images, as the network was trained on these.
The results can be used for the infrared images on the right as well, as the bar positions are identical in both image types.
The two strips of Japanese paper, or hinges, used to hold the fragment in place can also be seen at the top.
Additional paper was used by conservators to reinforce the verso, and is visible through a hole in the parchment near the top of the fragment. Images courtesy of Leon Levy Dead Sea Scrolls Digital Library, Israel Antiquities Authority; photo: Shai Halevi.}
\label{fig:bar_detection_result}
\end{figure}

The results of bar detection on some test set samples are visualized in Fig.~\ref{fig:bar_detection_result} with similar accuracy observed in the other test results as well.
These results support the quantitative evaluation of the trained model on the 38-image test set, where it achieved an Average Precision (AP) score of $92\%$.
The AP measures the average precision at different intersection over union (IoU) thresholds from $0.50$ to $0.95$ with a step size of $0.05$. The low AP score (92\%) is attributed to the small size of the train set. Despite this, the results are considered sufficient for the purposes of bar detection in the pipeline as high accuracy is not necessarily required in this step.

\subsection{Align Images of Recto and Verso}

The alignment of the image of a fragment recto with an image of its verso enables us most importantly to avoid interpreting ink marks on the recto as holes, since the ink will be black on the recto but will not be detectable on the image of the fragment verso, whereas a hole will be visible on both the recto and verso.
Two caveats apply here: A minority of fragments are opisthographs, written on both sides \cite{Perrot+2020+101+114}, in which case ink may occur at the same location on both sides of the fragment and will be interpreted here as a hole; an imaged fragment may be a ``wad'', a stack of two or more fragments that cannot be safely separated from each other \cite{Chapter7FindingWads}, with the result that the recto/verso image set depicts the recto of the surface of the fragments visible from the top of the stack and the verso shows only those visible from the bottom.
In the second case, the approach here cannot be used to reliably isolate any single fragment in the wad, further progress would require either a conservationist to separate individual fragments of the wad or a different type of imaging or detection technology to isolate the individual items in the stack.

The alignment of recto/verso sets is performed on the infrared images of the fragment.
First, the bars were masked using the bar detection model described in Section \ref{sec:ruler}.
The verso image was then horizontally flipped.
Next, we carried out keypoint detection and description on the two.
Keypoints are regions with maximum variance when moved around it, and their descriptors are vectors that describe the regions surrounding them.
We then take the descriptor of one keypoint in recto and match it with all other keypoints in verso to find the closest two keypoints, so that we can apply a ratio test \cite{lowe2004distinctive} that  discards many of the false matches arising from background clutter,
because, for false matches, there will likely be a number of other false matches within similar distances.
We reject all matches for which the distance ratio is greater than $0.80$.
This does not remove matches from other valid regions.
To solve this problem, we fit the alignment matrix using  RANSAC (Random Sample Consensus) method that iteratively fits the alignment matrix by selecting a random subset of the keypoints and refining the model based on that subset until the best model is obtained that fits the majority of the data points within a specified error tolerance.
The error tolerance in RANSAC is a predefined threshold value that determines how far a data point can be from the model.

The final alignment's accuracy was evaluated based on the quantity of inliers in the final set of keypoint matches.
To enhance the outcome, the algorithm was executed using four different feature extractors (SIFT \cite{lowe2004distinctive}, KAZE \cite{alcantarilla2012kaze}, AKAZE \cite{alcantarilla2011fast}, and ORB \cite{rublee2011orb}) and with a threshold range from $21$ to $5$, decremented by $2$ at each step.
Analysis revealed that the optimal combination was obtained using SIFT with a threshold of $15$, resulting in the greatest number of inliers.

\begin{figure}[t]  %% better on top of page not in middle
 \centering \includegraphics[draft=false,width=\textwidth]{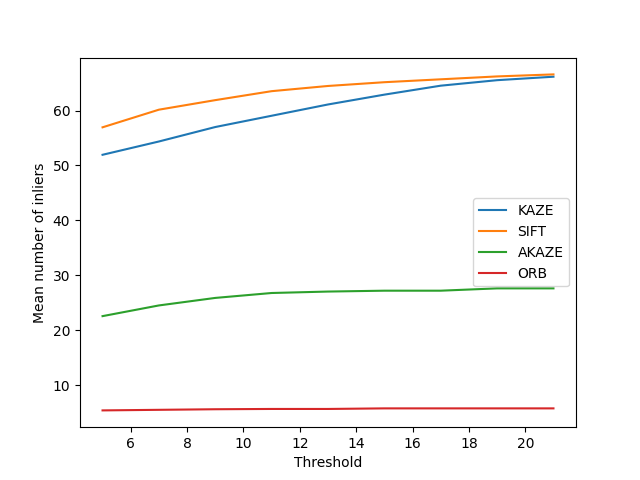} %[scale=1.7]
  \caption{The $y$-axis represents the average number of inliers across the test images, while the $x$-axis represents the threshold value. Each plot line represents the results for a different feature extractor, with the highest number of inliers being returned using SIFT with a threshold value of $15$.}
  \label{fig:inliers}
\end{figure}

\begin{figure}[t]  %% better on top of page not in middle
 \centering \includegraphics[draft=false,width=0.9\linewidth]{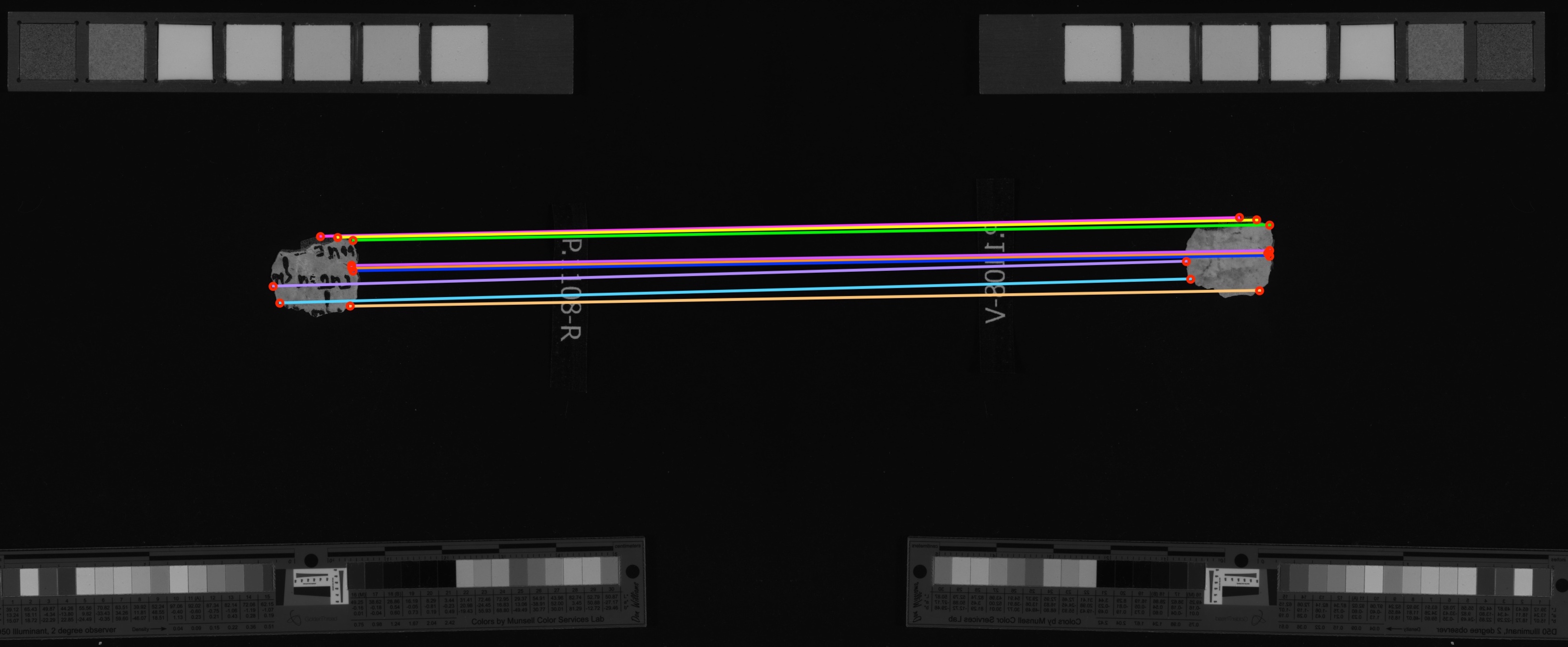} %[scale=1.7]
  \caption{Recto and flipped verso infrared images, with inliers depicted as red circles and their corresponding match connections as colored lines. Images courtesy of Leon Levy Dead Sea Scrolls Digital Library, Israel Antiquities Authority; photo: Shai Halevi.}
  \label{fig:inlier_matches}
\end{figure}

\subsection{Image Thresholding}

\begin{figure}[t]
  \centering
\begin{tabular}{c c c}
& \textbf{Original} & \textbf{Thresholded} \\
\rotatebox{90}{\textbf{Recto}} & \includegraphics[draft=false,width=0.48\textwidth]{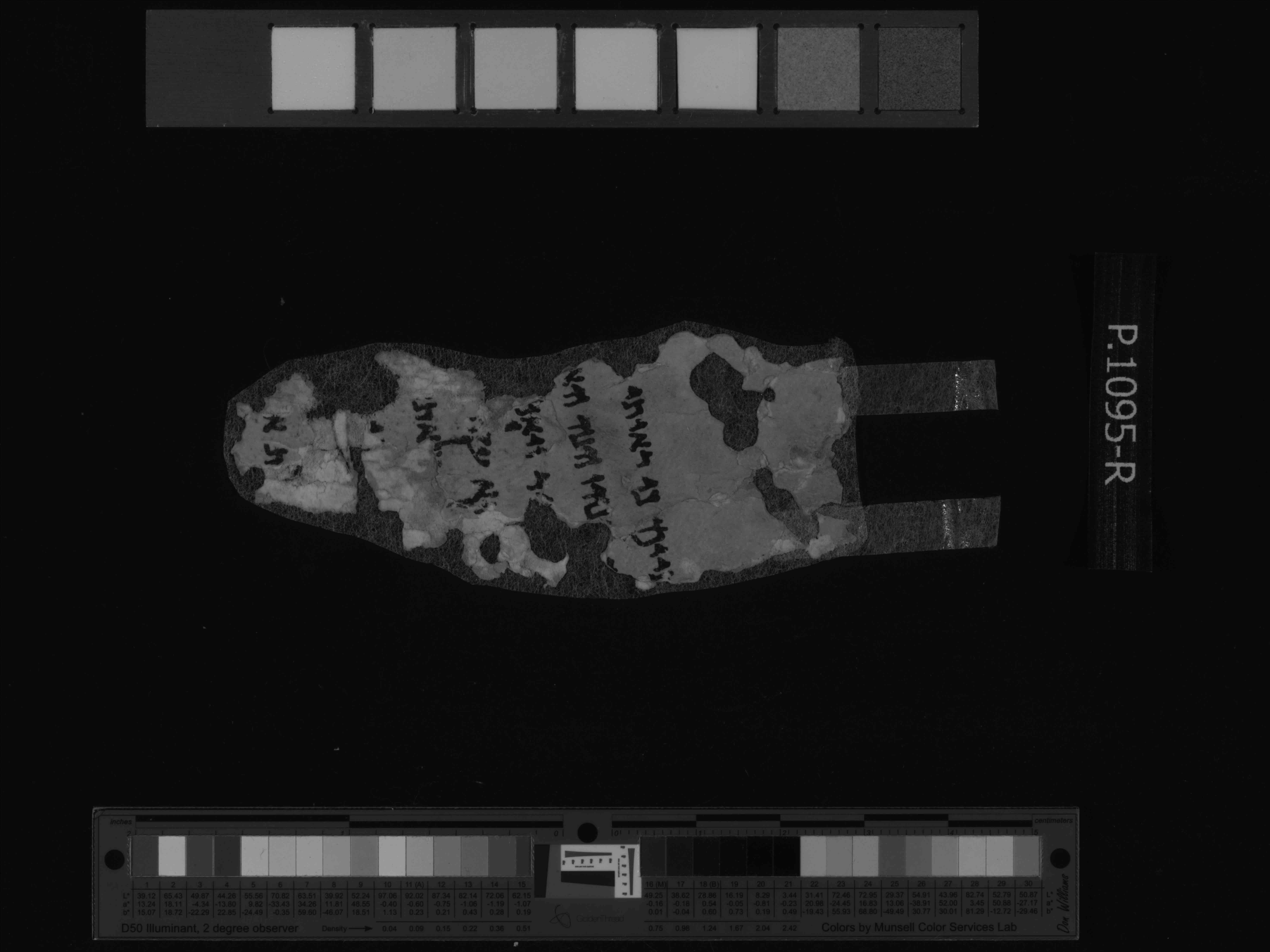} & \includegraphics[draft=false,width=0.48\textwidth]{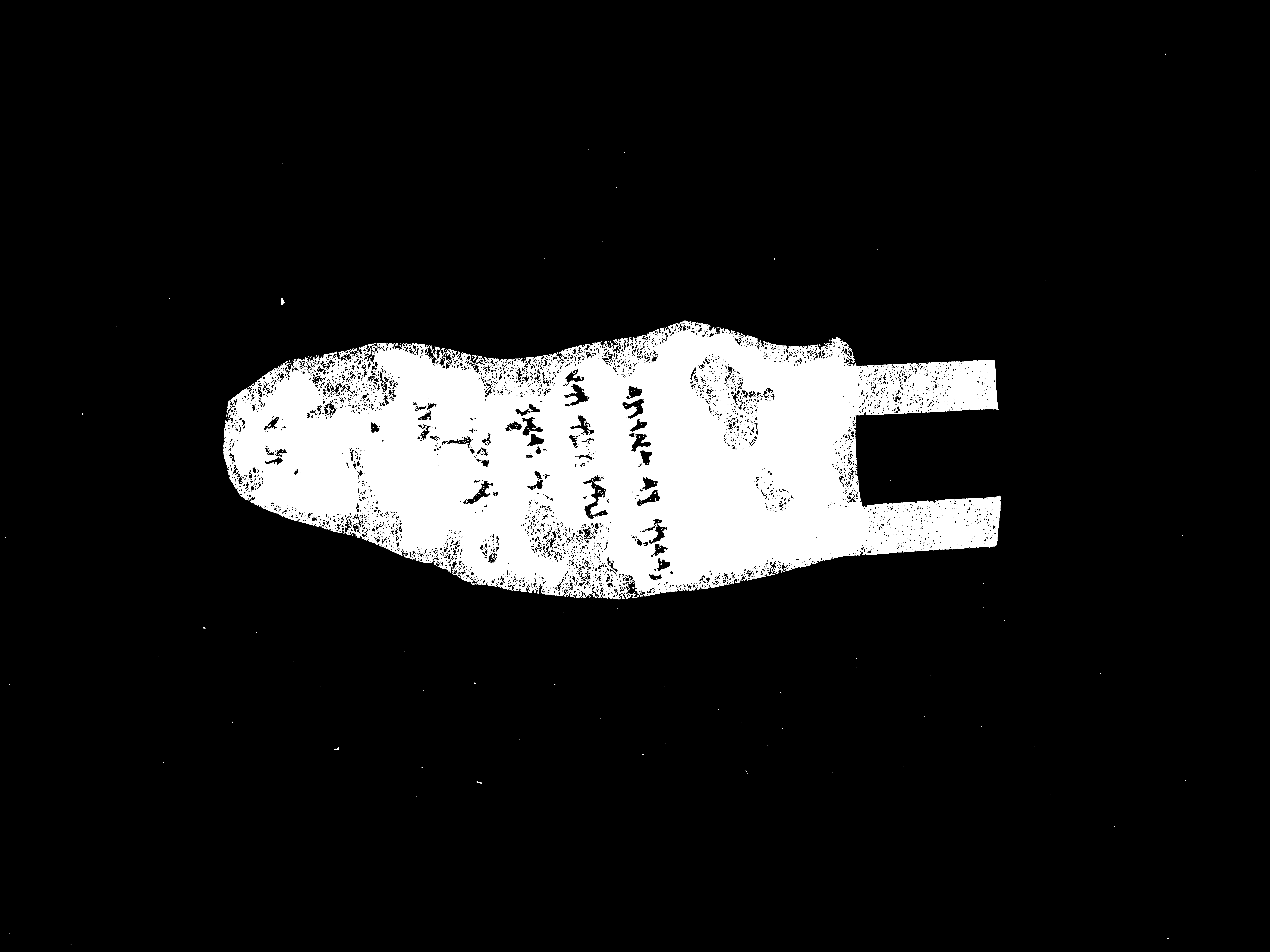} \\
\rotatebox{90}{\textbf{Verso}} & \includegraphics[draft=false,width=0.48\textwidth]{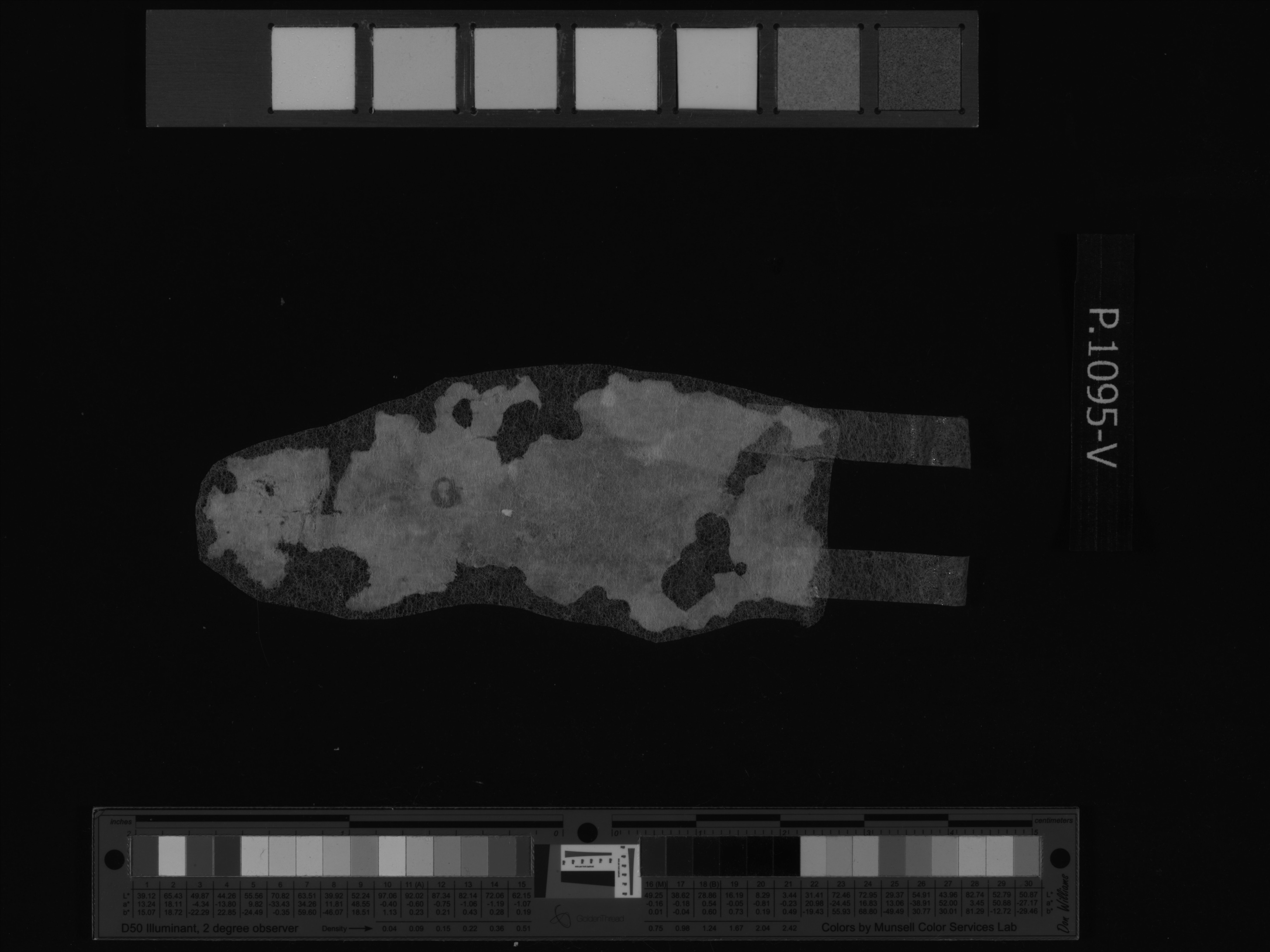} & \includegraphics[draft=false,width=0.48\textwidth]{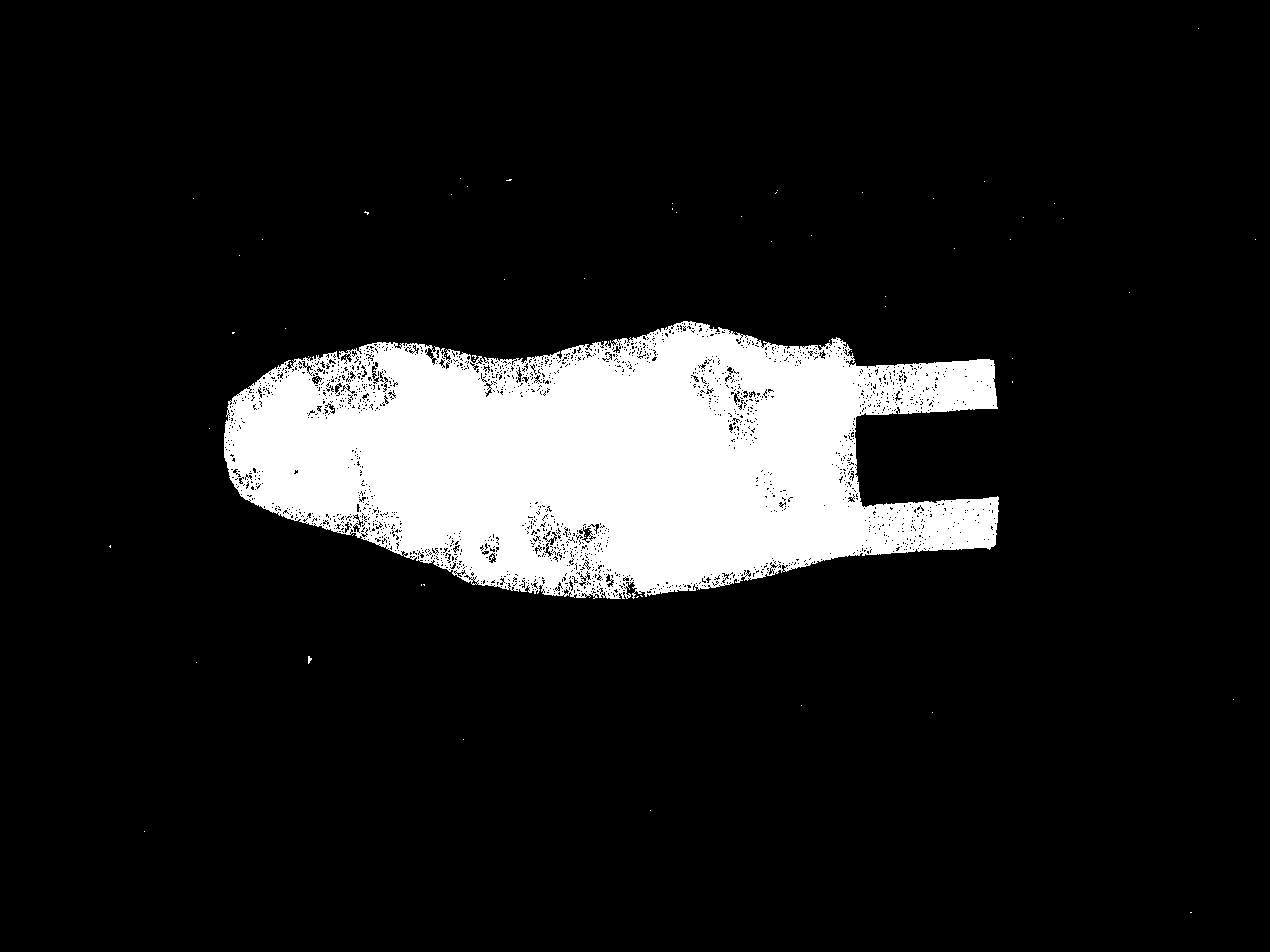} \\
\end{tabular}
\caption{Image Thresholding. Images courtesy of Leon Levy Dead Sea Scrolls Digital Library, Israel Antiquities Authority; photo: Shai Halevi.}
  \label{fig:thresholding}
\end{figure}

We preferred infrared images instead of color images because they have a higher contrast between the fragment and the background. In contrast, color images may have variations in lighting, shading, and color that can make thresholding a more challenging task. We have chosen to perform the first threshold using a dynamically calculated value. This is performed on the 8 bit grayscale infrared images by finding the average of all pixels with a value below 50 (i.e., all very dark pixels), to which a buffer of 10 is added. This tends to perform very well as a floor for removing the black background in our image set while leaving even very dark fragments or portions of fragments that have darkened due to age and exposure to the elements. This step provides a maximal fragment mask for the image of the fragment's recto and the image of the fragment's verso (Fig.~\ref{fig:thresholding}).

\subsection{First Mask}

\begin{figure}[t]
  \centering
\begin{tabular}{c c c}
\textbf{Recto} & \textbf{Verso} & \textbf{Combined} \\
\includegraphics[draft=false,width=0.32\textwidth]{figures/thresholding/1095_2_recto_pre.png} & \includegraphics[draft=false,width=0.32\textwidth]{figures/thresholding/1095_2_verso_pre.png} & \includegraphics[draft=false,width=0.32\textwidth]{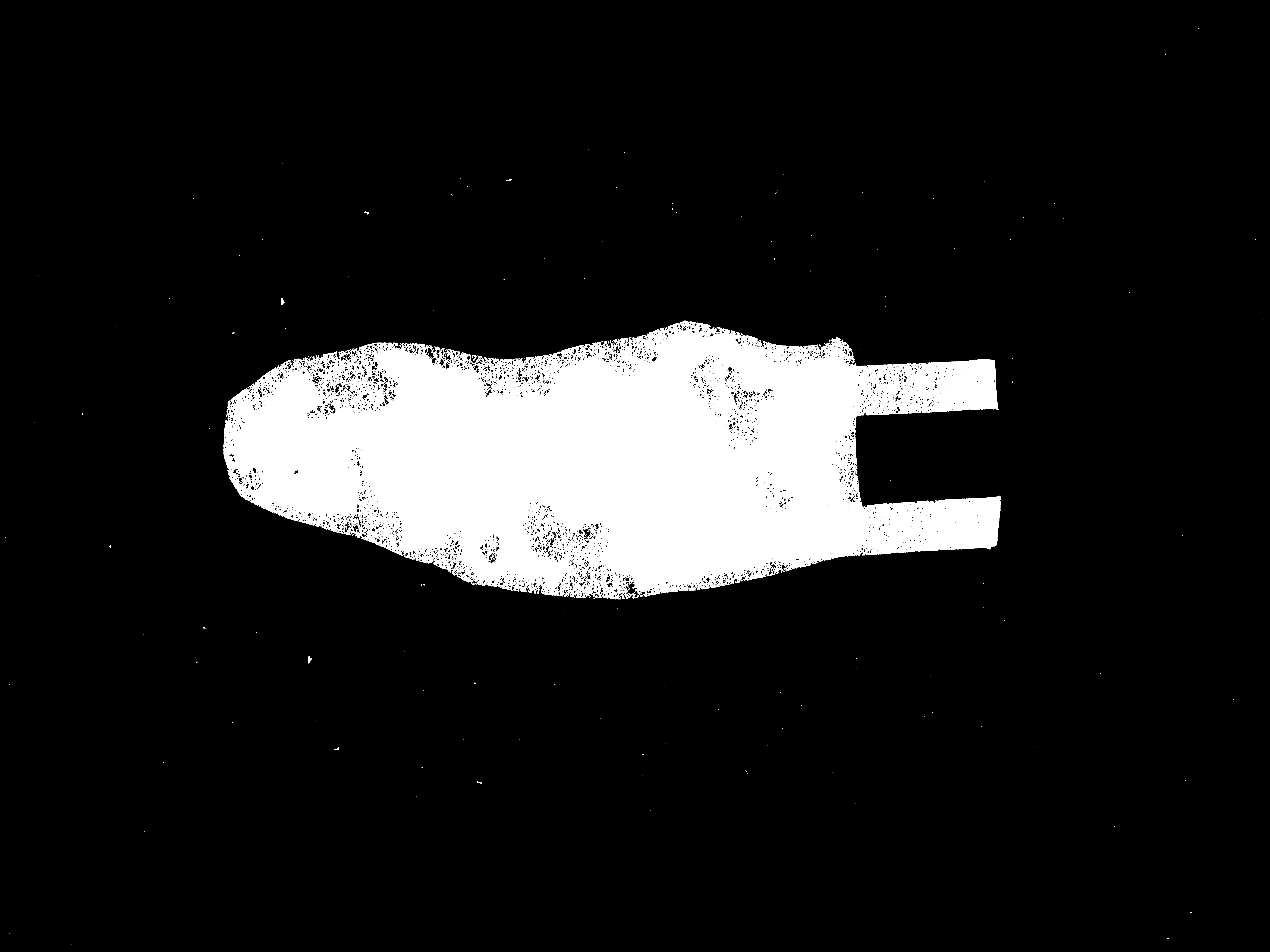} \\
\end{tabular}
\caption{First mask.}
  \label{fig:first_mask}
\end{figure}

Since we already have the transform matrix needed to match the image of the recto to the image of the verso, we apply this matrix to the verso mask in order to align the two masks from step 3. The recto and verso masks are then combined according to their union resulting in a maximal mask for the fragment itself. The separate recto and verso masks are retained for further refinements in step 8 (Fig.~\ref{fig:first_mask}).

\subsection{Masking the Backing Substrate}

\begin{figure}[t]
  \centering
\begin{tabular}{c c c}
& \textbf{Original} & \textbf{Backing Mask} \\
\rotatebox{90}{\textbf{Recto}} & \includegraphics[draft=false,width=0.48\textwidth]{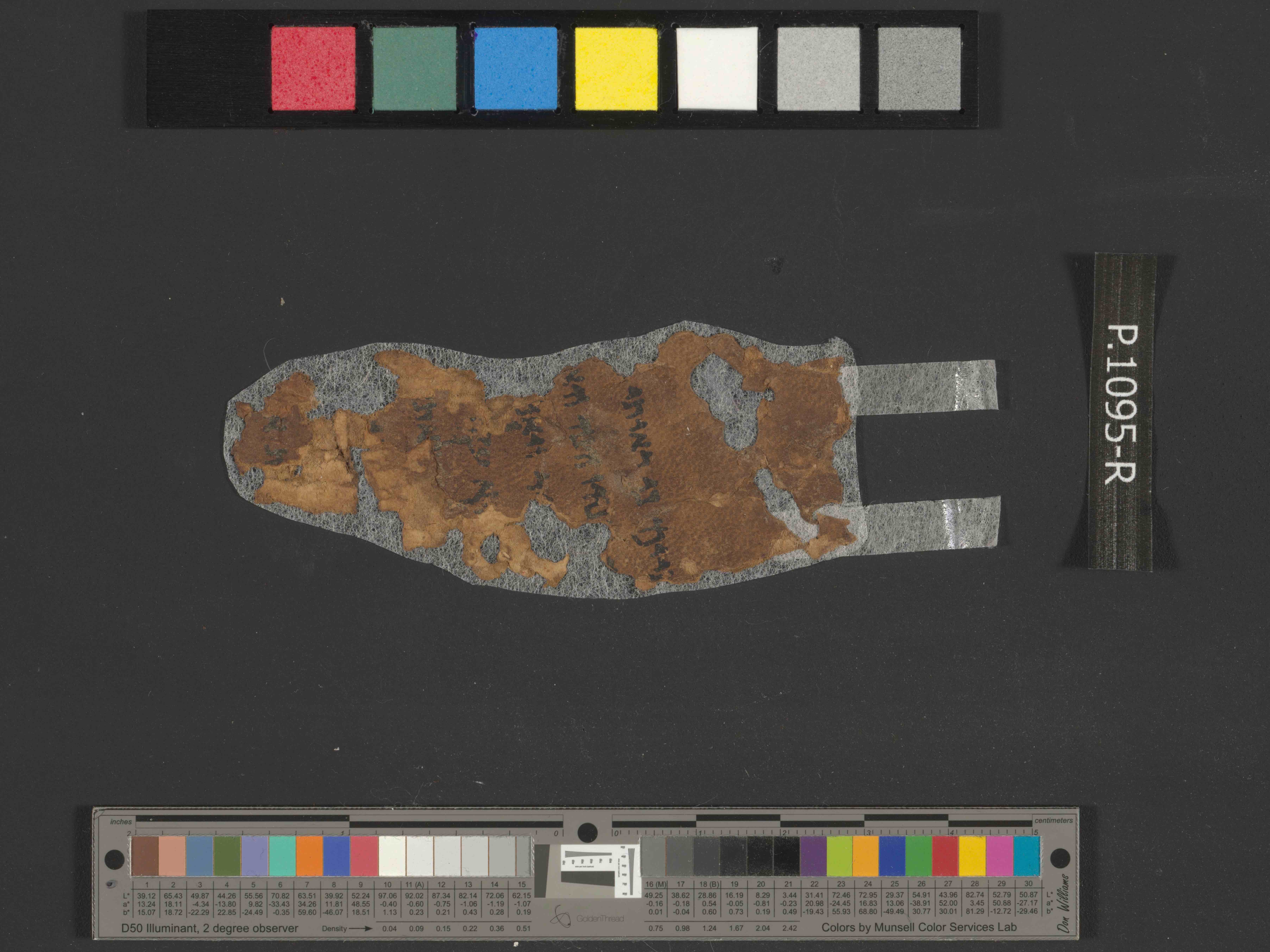} & \includegraphics[draft=false,width=0.48\textwidth]{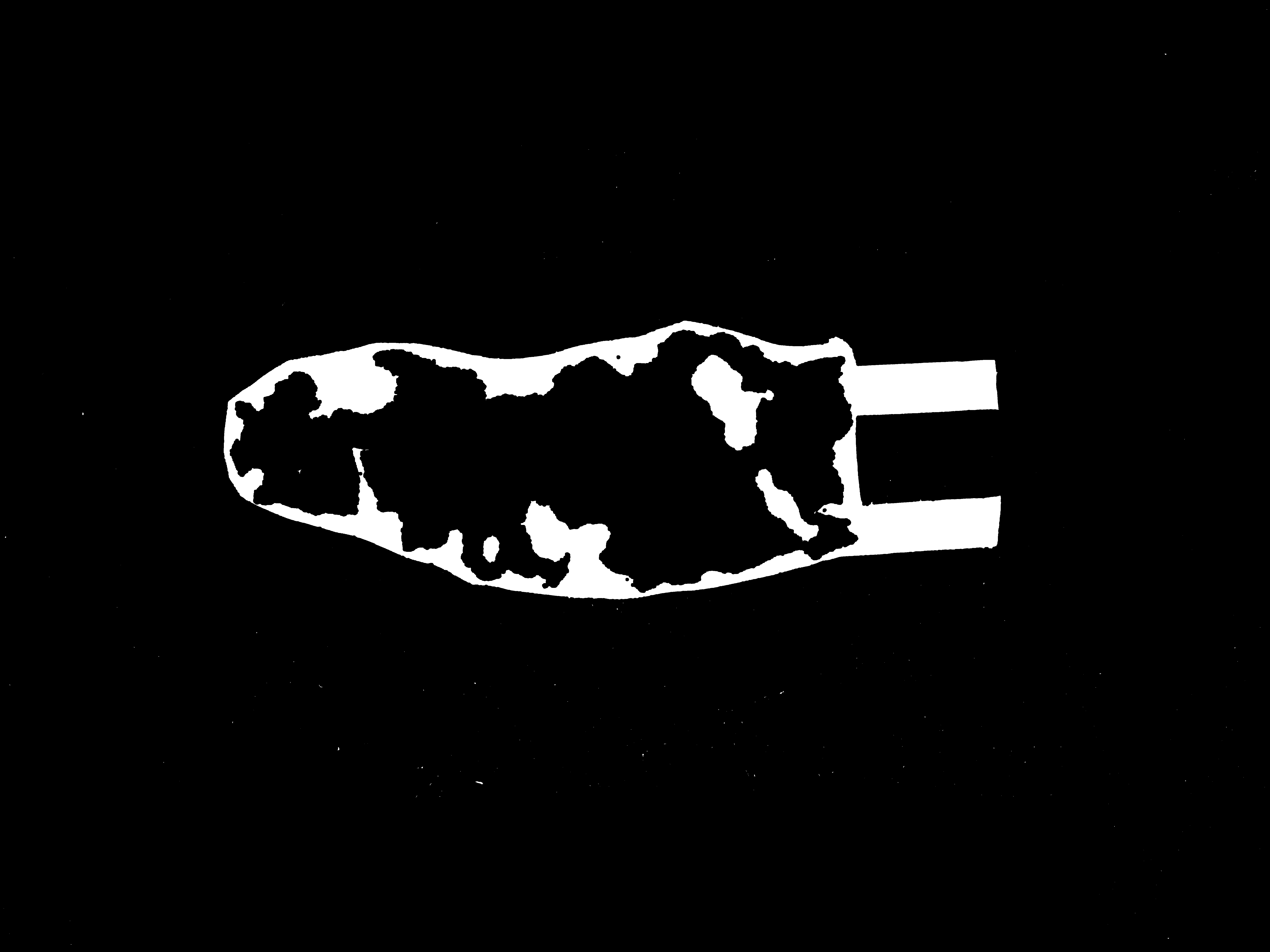} \\
\rotatebox{90}{\textbf{Verso}} & \includegraphics[draft=false,width=0.48\textwidth]{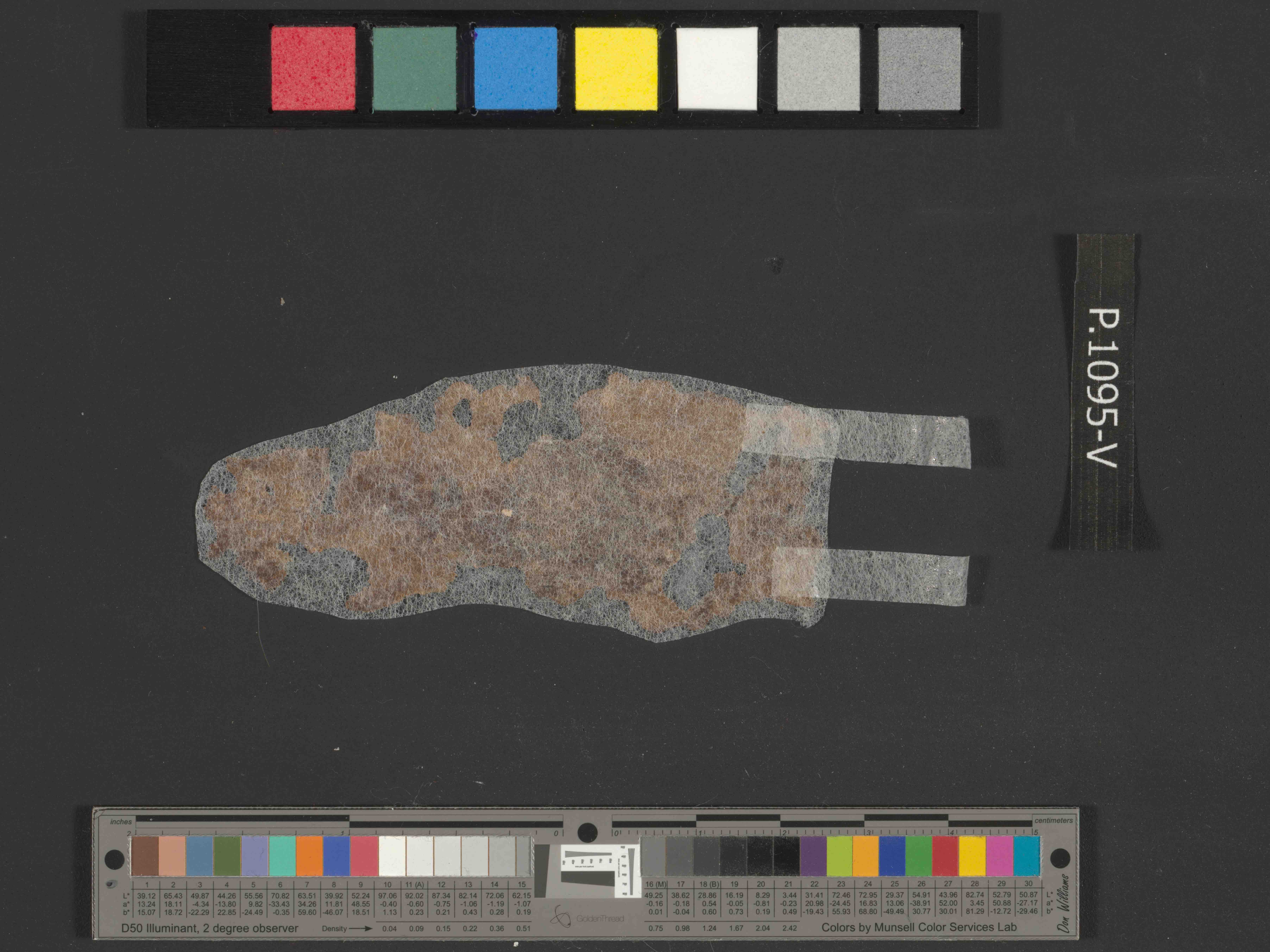} & \includegraphics[draft=false,width=0.48\textwidth]{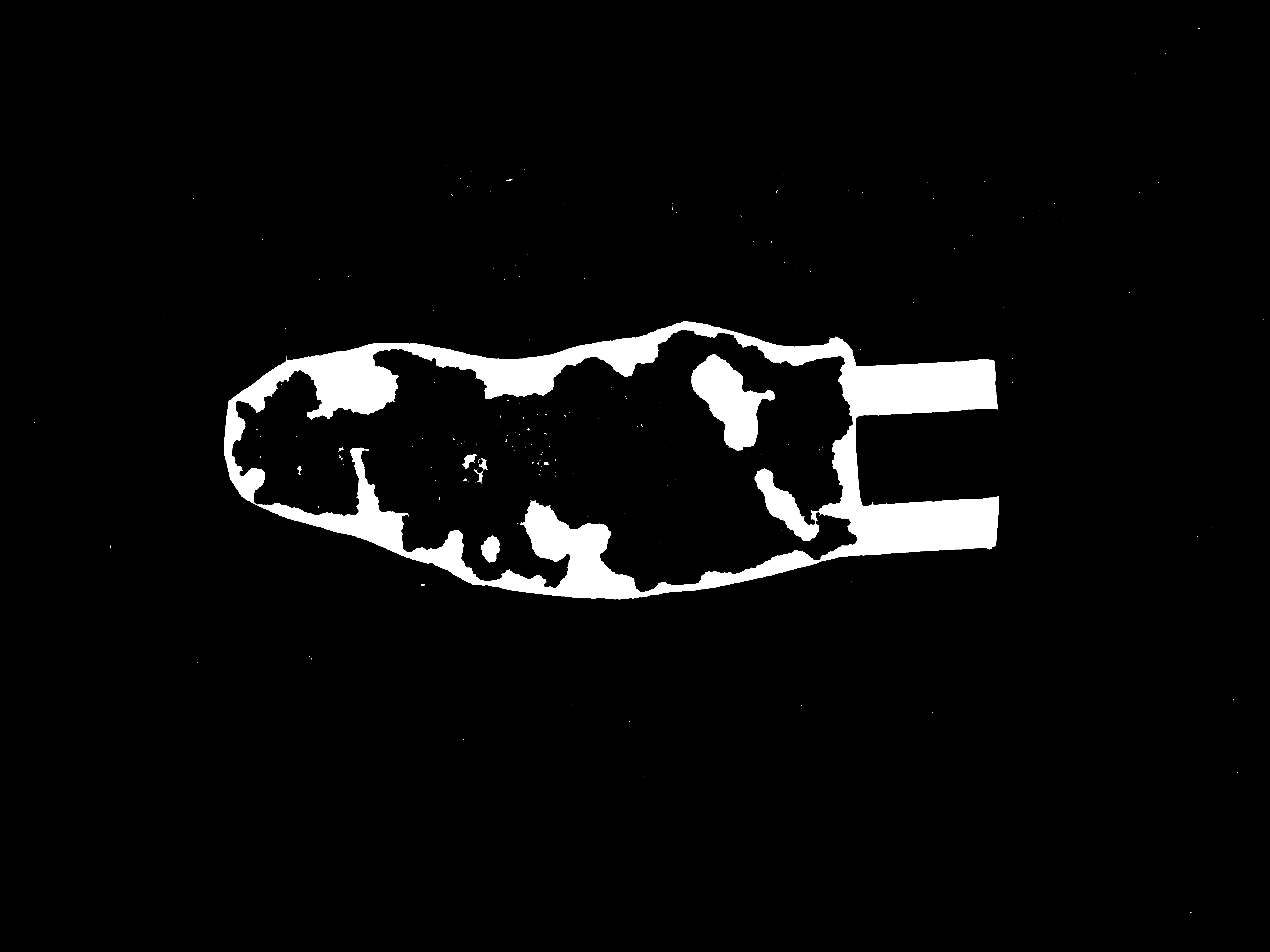} \\
\end{tabular}
\caption{Backing substrate masks. Images courtesy of Leon Levy Dead Sea Scrolls Digital Library, Israel Antiquities Authority; photo: Shai Halevi.}
  \label{fig:backing_mask}
\end{figure}

The backing substrate used by the conservationists at the IAA is highly detectable in the full-color images when they are transformed into full HSV color-space.
For our image set, we used manual testing to determine that backing substrate occurs with HSV values between (0, 0, 100) and (255, 20, 200). We create masks for the image of the recto and the image of the verso using all pixels that fall within that range (Fig.~\ref{fig:backing_mask}).

\paragraph*{Morphological closing.}
The resulting masks tends to have many small holes due to the woven nature of the substrate. They are improved by first discarding any tiny elements and then performing a morphological close operation. The IAA images all have a similar pixel density very close to 1215 PPI and a 20$\times$20 pixel ellipse-shaped kernel appears to provide the optimal results for closing.

\begin{figure}[t]
  \centering
\begin{tabular}{c c c}
\textbf{Recto} & \textbf{Verso} & \textbf{Combined} \\
\includegraphics[draft=false,width=0.32\textwidth]{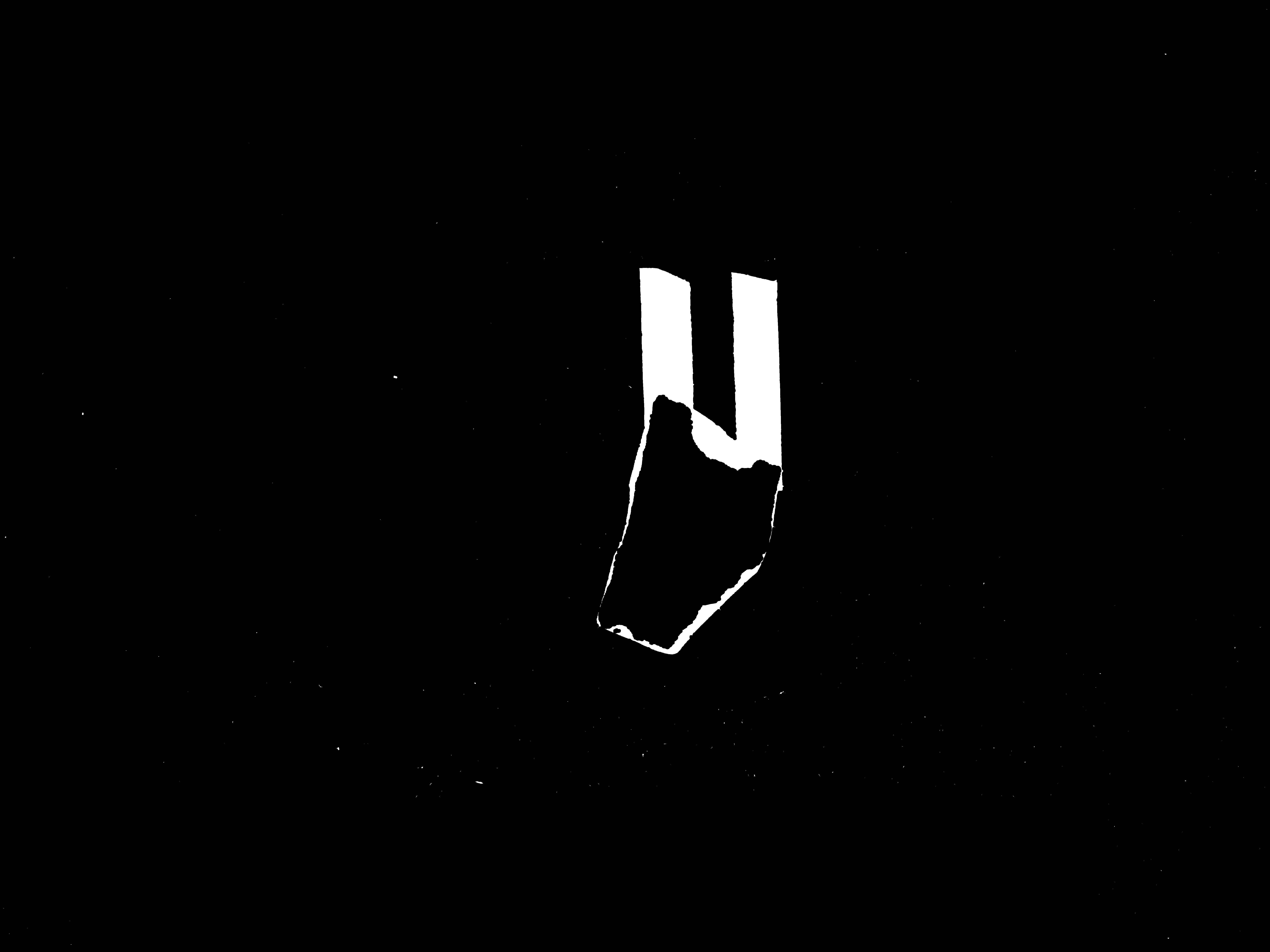} & \includegraphics[draft=false,width=0.32\textwidth]{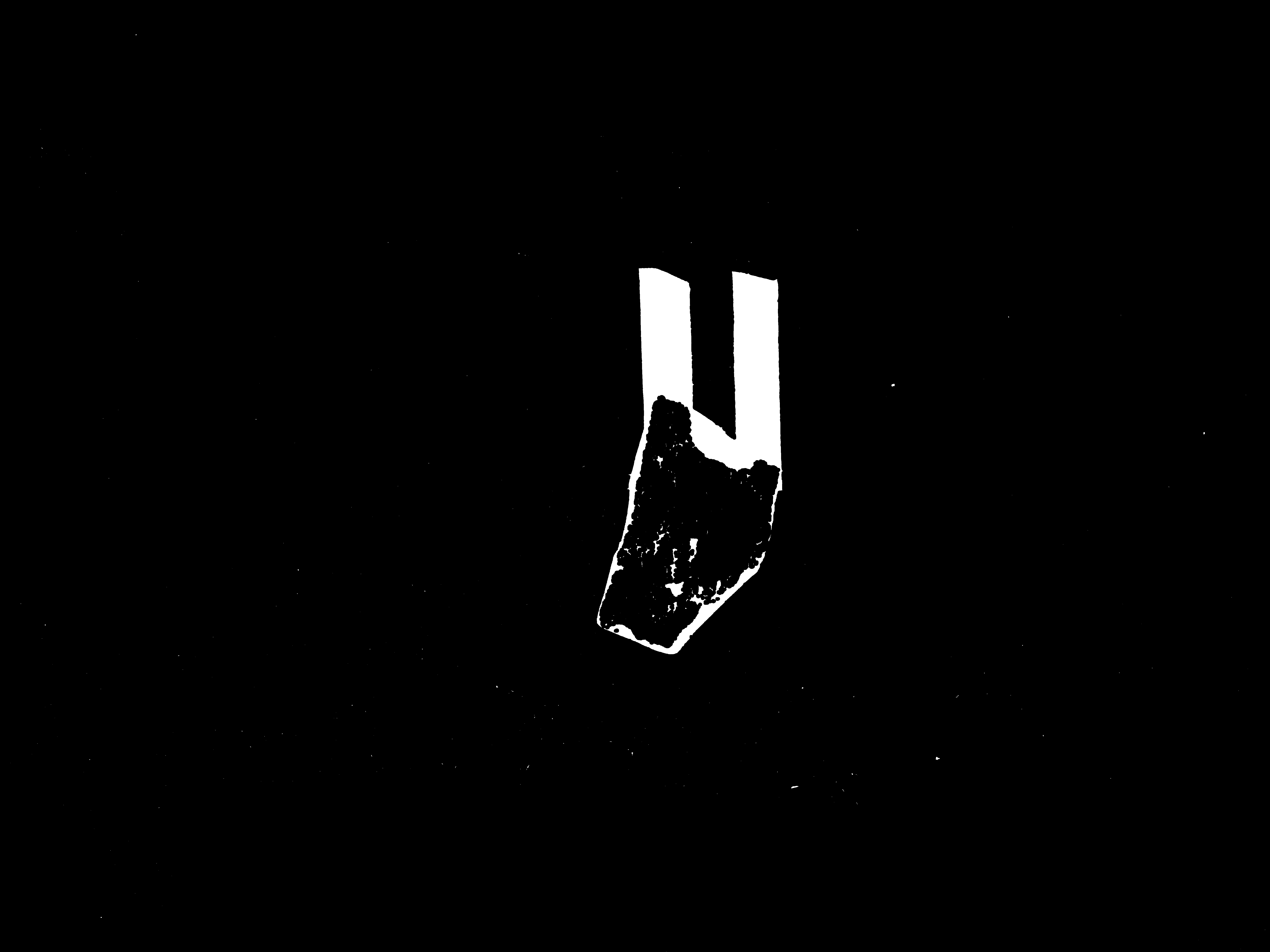} & \includegraphics[draft=false,width=0.32\textwidth]{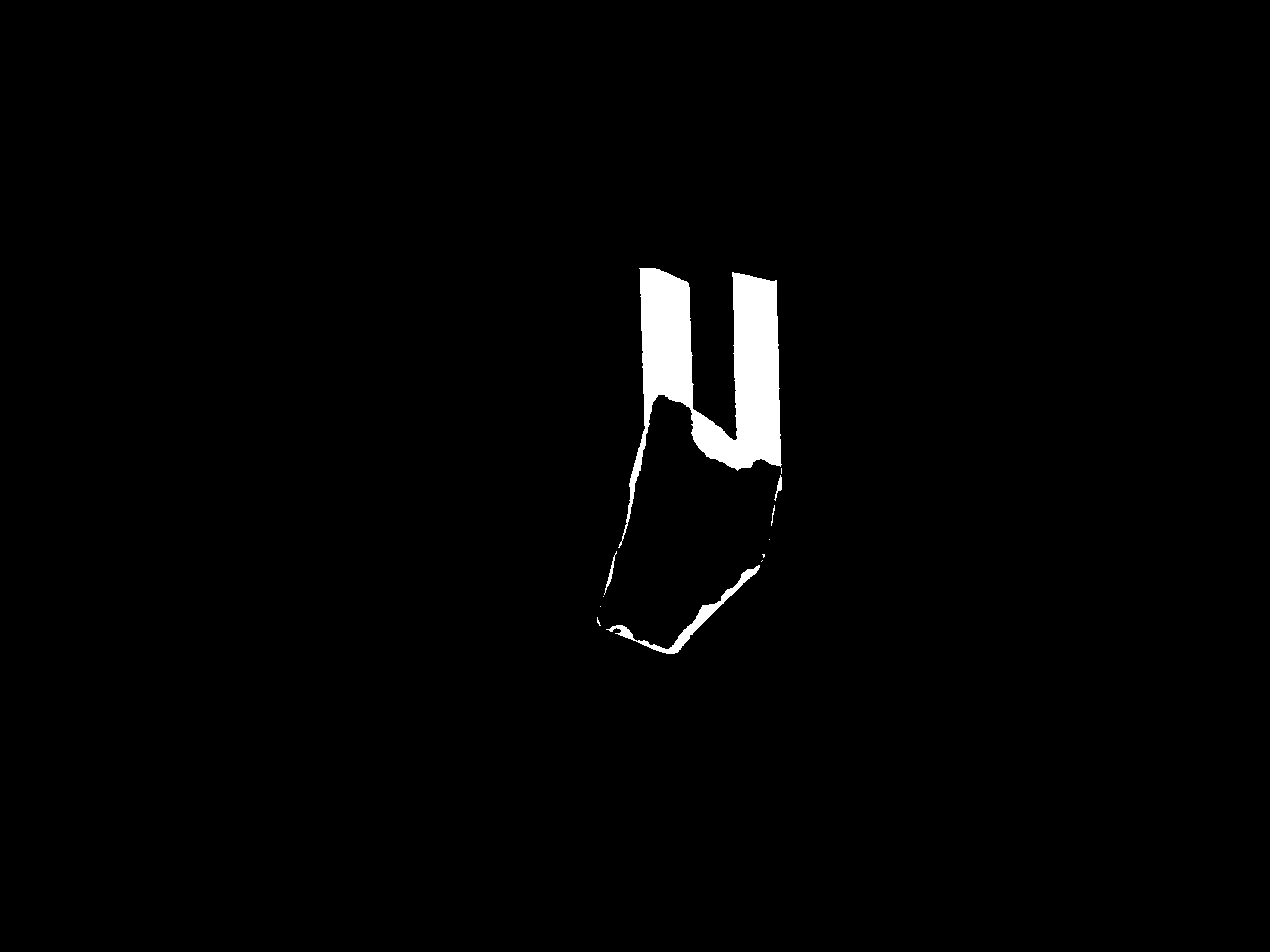} \\
\end{tabular}
\caption{Combination of backing masks.}
  \label{fig:combined_backing_mask}
\end{figure}

Often more of the backing substrate will be visible on the verso of the fragment than the recto, in fact the entire verso of the fragment may be covered by the backing substrate. For this reason, we use only the intersection of the recto and verso masks. This results in a mask for only those portions of the backing substrate that do not overlap with the manuscript fragment or fragments (Fig.~\ref{fig:combined_backing_mask}).

\subsection{Preparing the Final Mask}

\begin{figure}[t]
  \centering
\begin{tabular}{c c c c}
\textbf{Initial Recto} & \textbf{Initial Verso} & \textbf{Backing Mask} &\textbf{Final Mask} \\
\includegraphics[draft=false,width=0.23\textwidth]{figures/thresholding/1095_2_recto_pre.png} & \includegraphics[draft=false,width=0.23\textwidth]{figures/thresholding/1095_2_verso_pre.png} & \includegraphics[draft=false,width=0.23\textwidth]{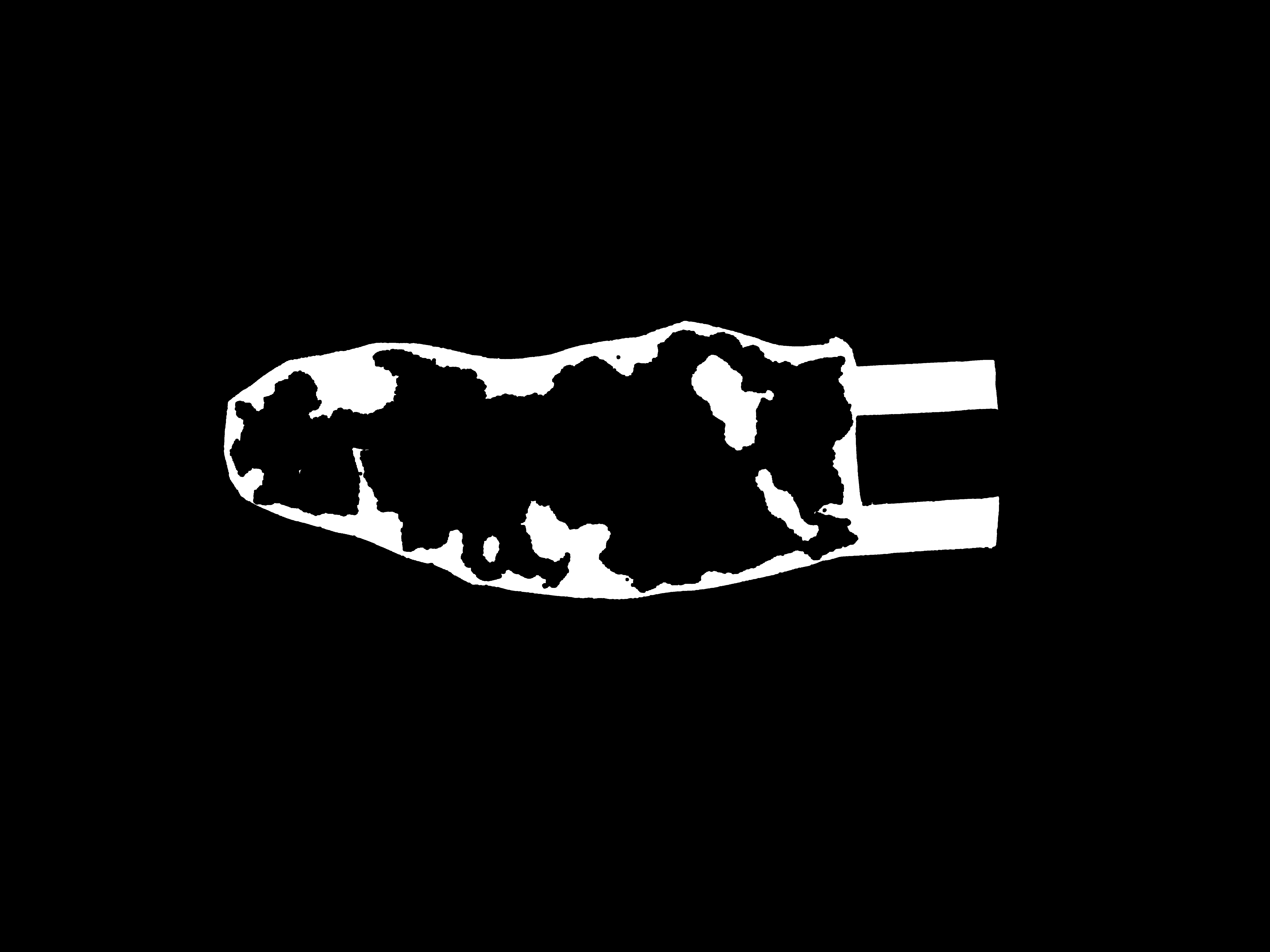} & \includegraphics[draft=false,width=0.23\textwidth]{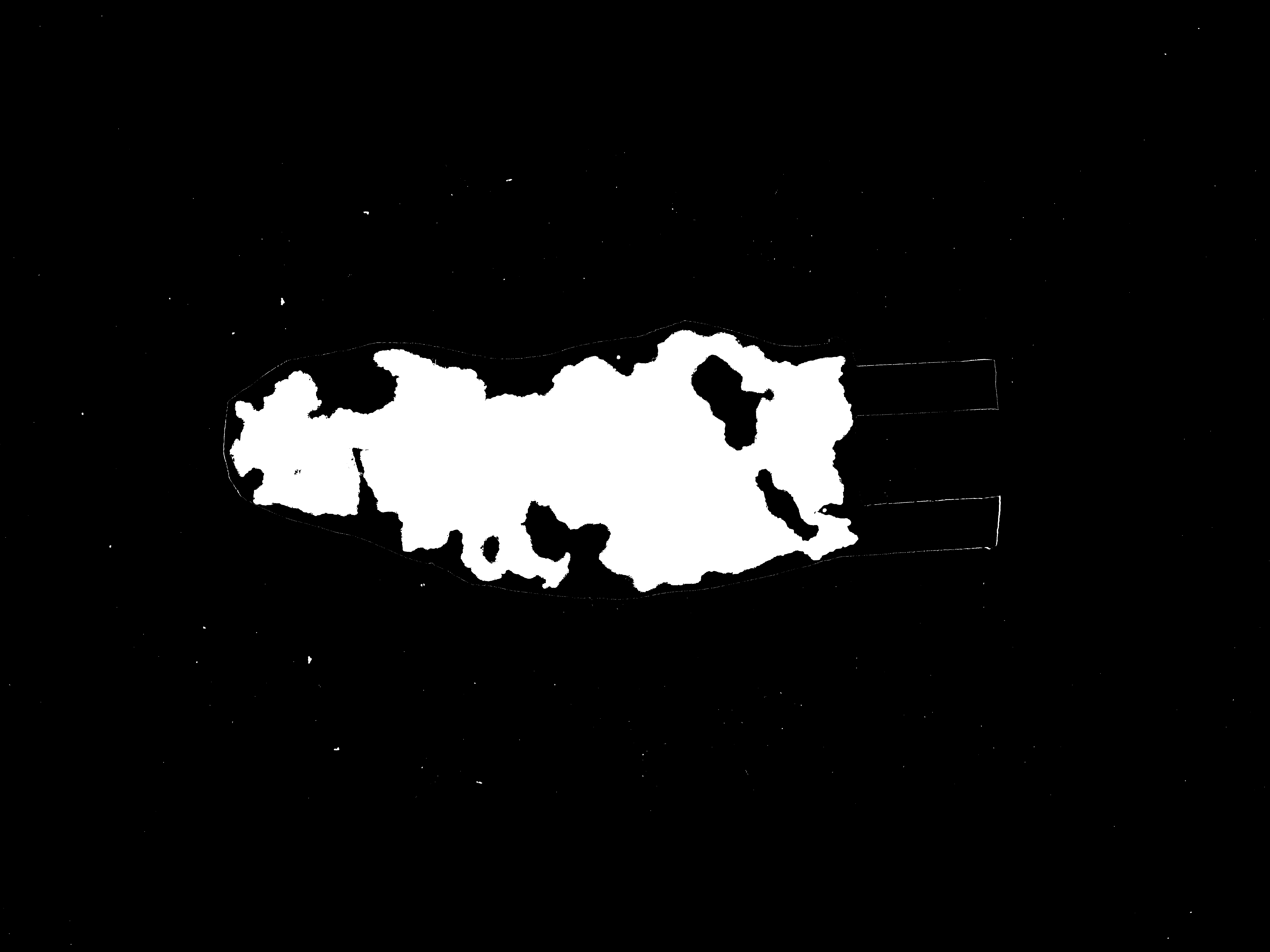} \\
\end{tabular}
\caption{Final mask.}
  \label{fig:final_mask}
\end{figure}

The backing material mask from step 5 is subtracted from the fragment mask in step 4. The result is a mask that should cover all of the fragment or fragments in the images excluding any portions that only consist of the backing material (Fig.~\ref{fig:final_mask}).

\subsection{Calculating Fragment Locations}

We find the contours in the final mask and sort them according to hierarchy into separate polygons with holes.
The vector shape or shapes produced here should mainly correspond to the actual fragment or fragments in the image set---there may be more than one manuscript fragment in an image set.
The contour detection algorithm in OpenCV may result in polygons that are invalid topographies according to the OGC standard,%
\footnote{\url{https://portal.ogc.org/files/?artifact_id=25355 6.1.11.1}.} which will create problems for some algorithms for (geo-) spatial analysis.
For this reason, the results  must be verified and repaired as best as possible.
For this we use the geo-repair-polygon library~\cite{geo-repair} that was created for this purpose within the Scripta Qumranica Electronica project.%
\footnote{DFG Projektnummer 282601852; \url{https://gepris.dfg.de/gepris/projekt/282601852}, \url{http://qumranica.org}.}

\subsection{Exclude Unwanted Fragments}

We check each fragment from step 7 against the recto and the verso masks from step 5.
Any fragment polygons that do not overlap with features in \emph{both} %\todo{italics are preferable for emphasis}
the recto and the verso masks from that step are discarded. This step removes any small features that are not directly attached to the fragments themselves (i.e., features that occur in different places in the recto and verso images).

\subsection{Final Filter}

\begin{figure}[t]
  \centering
\begin{tabular}{c c c}
\textbf{Highlighted Fragment} & \textbf{Fragment Vector} & \textbf{Extracted Fragment} \\
\includegraphics[draft=false,width=0.32\textwidth]{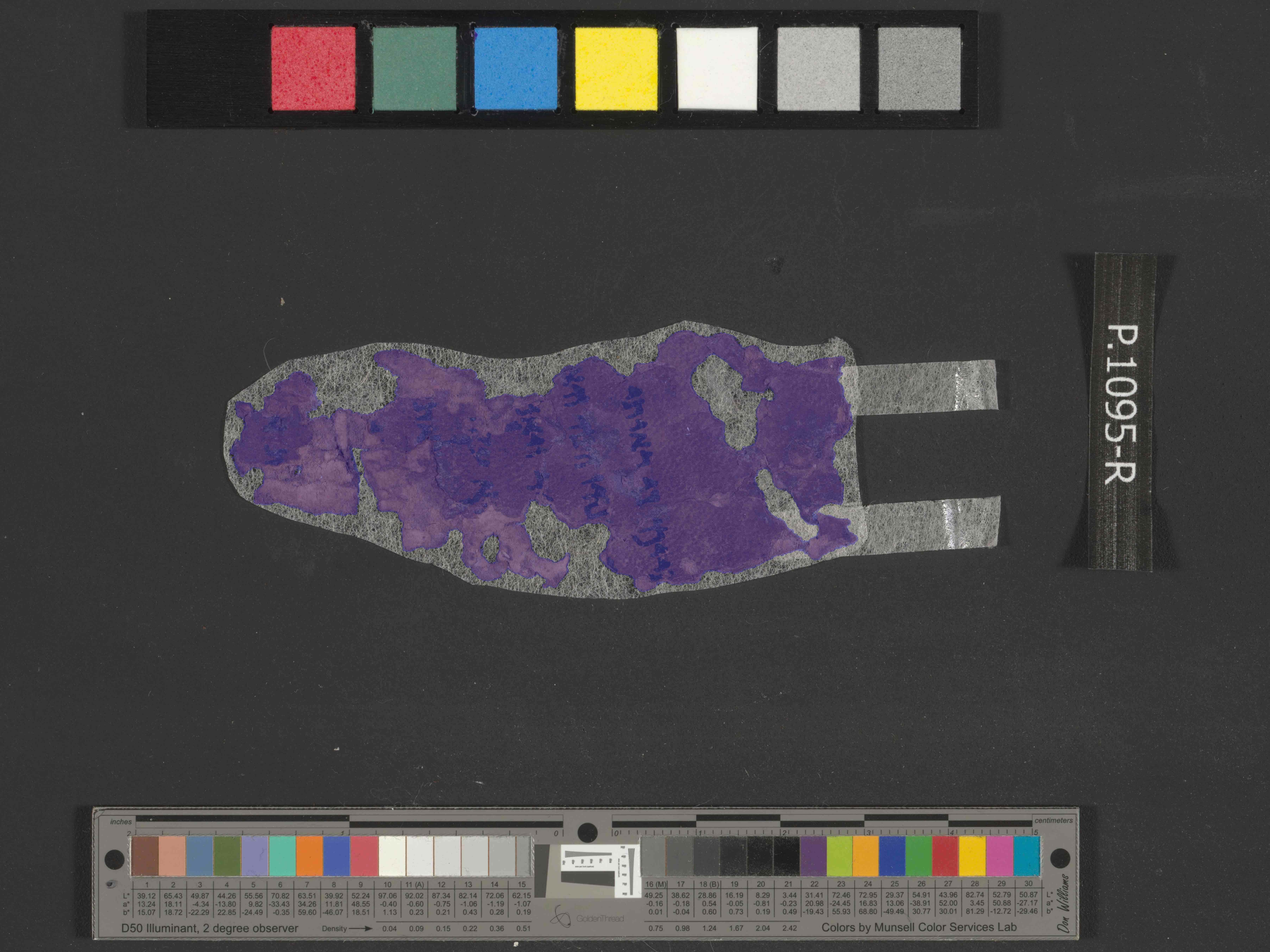} & \includegraphics[draft=false,width=0.32\textwidth]{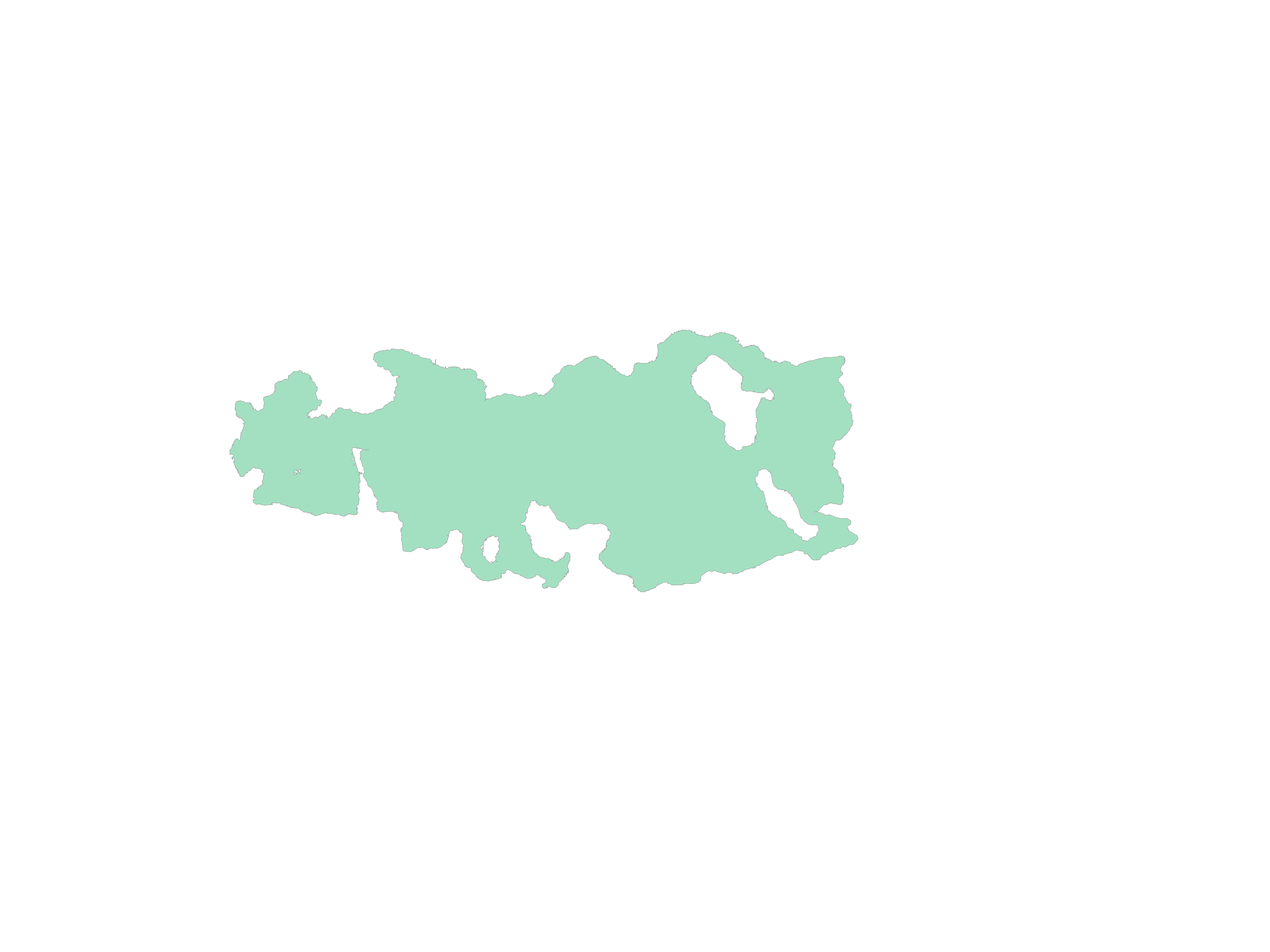} & \includegraphics[draft=false,width=0.32\textwidth]{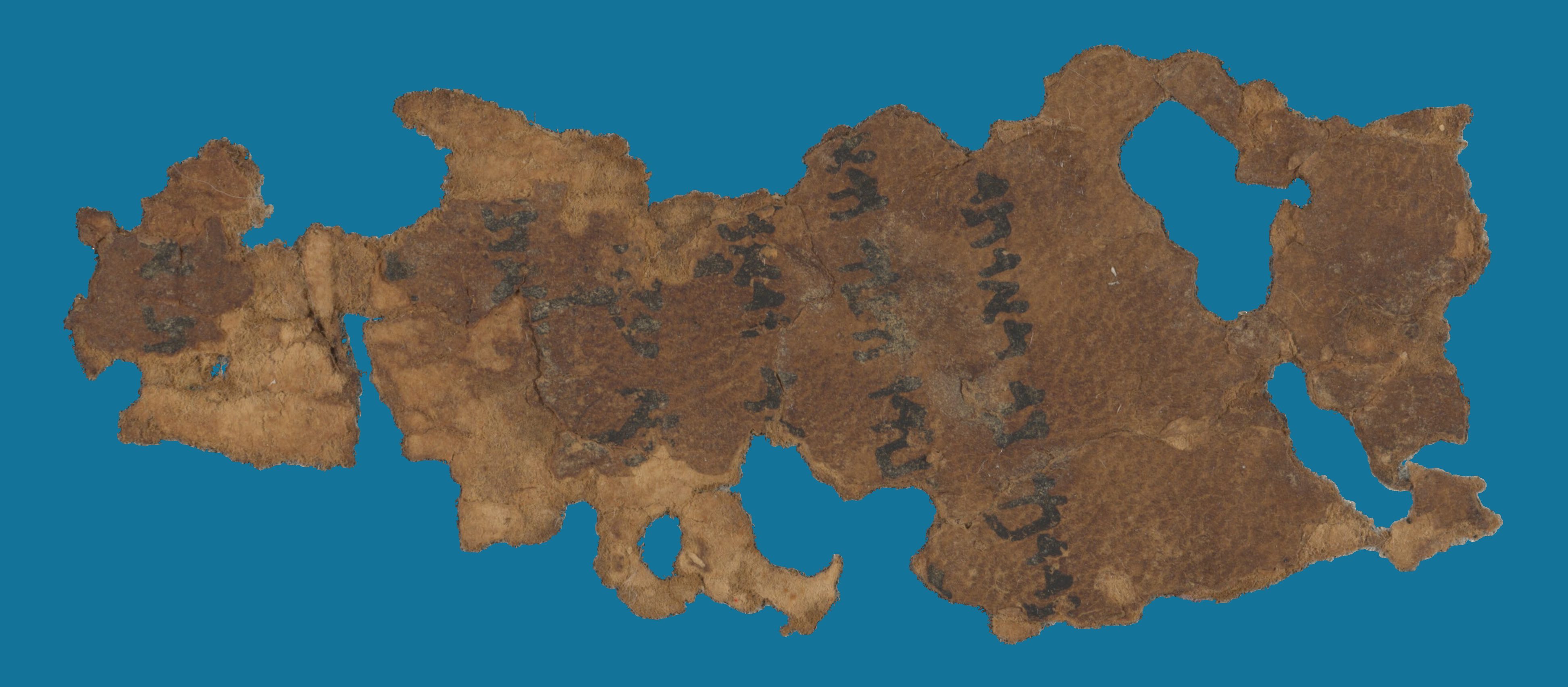} \\
\end{tabular}
\caption{Extracted fragment. Images courtesy of Leon Levy Dead Sea Scrolls Digital Library, Israel Antiquities Authority; photo: Shai Halevi.}
  \label{fig:final_fragment_extraction}
\end{figure}

\begin{figure}[t]
  \centering
\begin{tabular}{c c c c}
\multicolumn{1}{p{0.23\textwidth}}{\centering\textbf{ Highlighted Fragments}} & \textbf{Fragment 1} & \textbf{Fragment 2} & \textbf{Fragment 3} \\
\includegraphics[draft=false,width=0.23\textwidth]{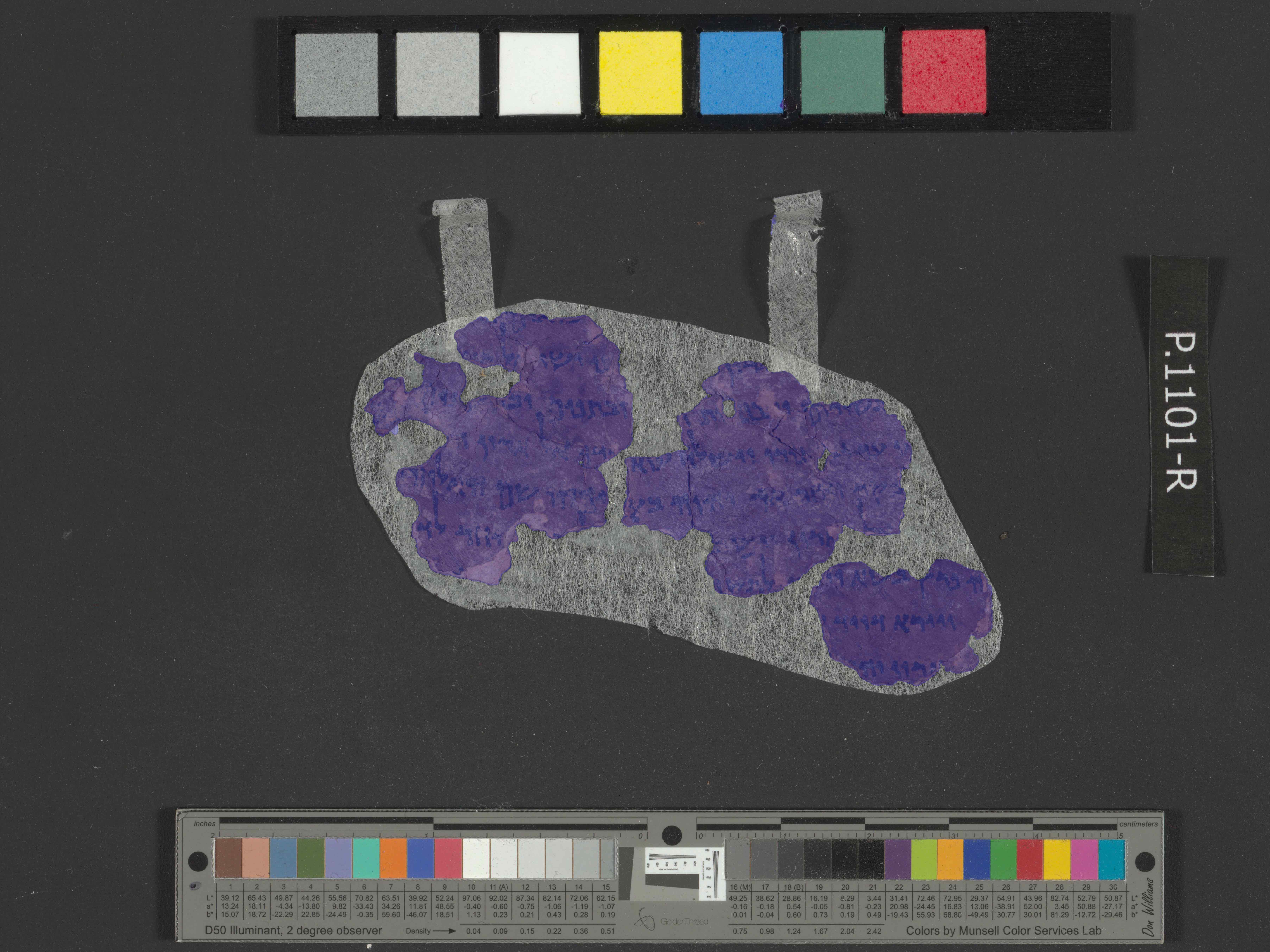} & \includegraphics[draft=false,width=0.23\textwidth]{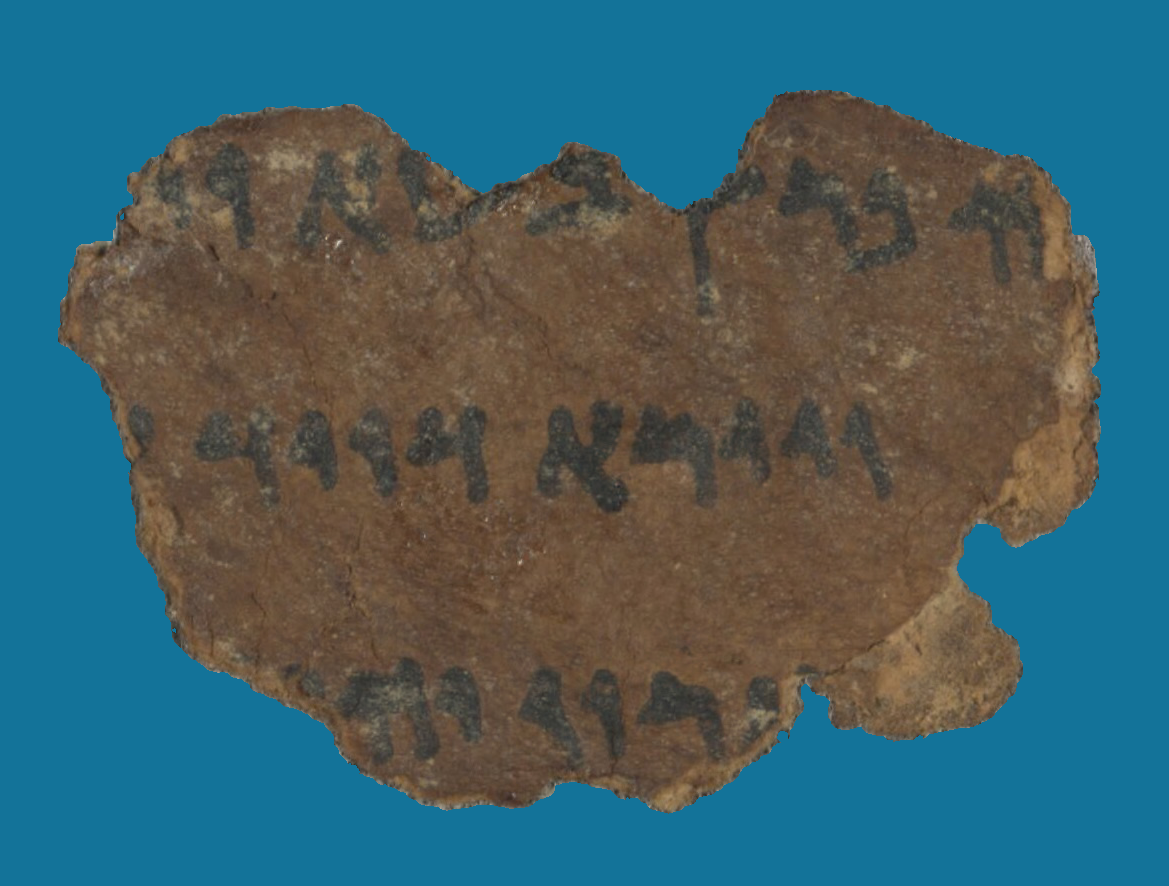} & \includegraphics[draft=false,width=0.23\textwidth]{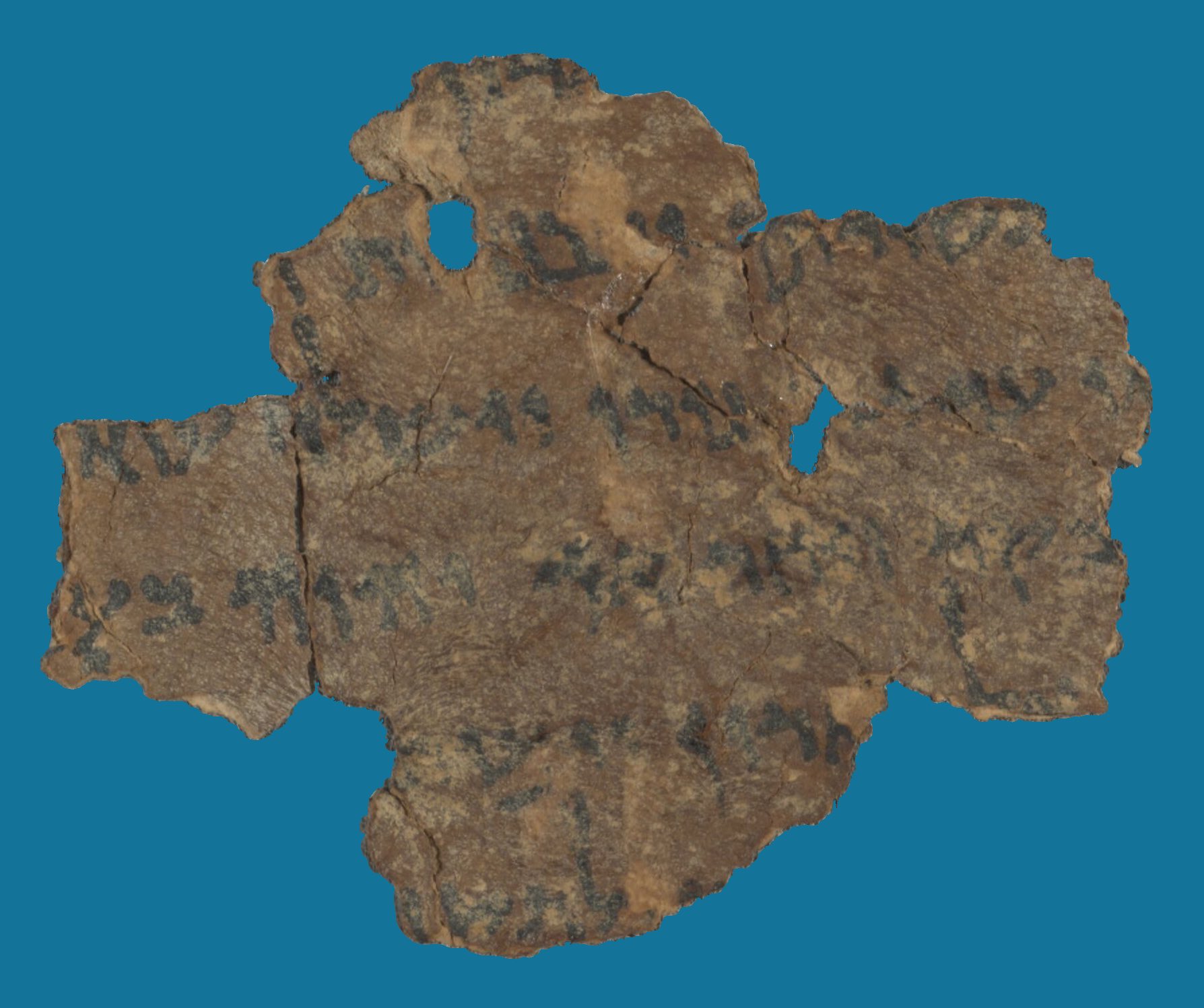} & \includegraphics[draft=false,width=0.23\textwidth]{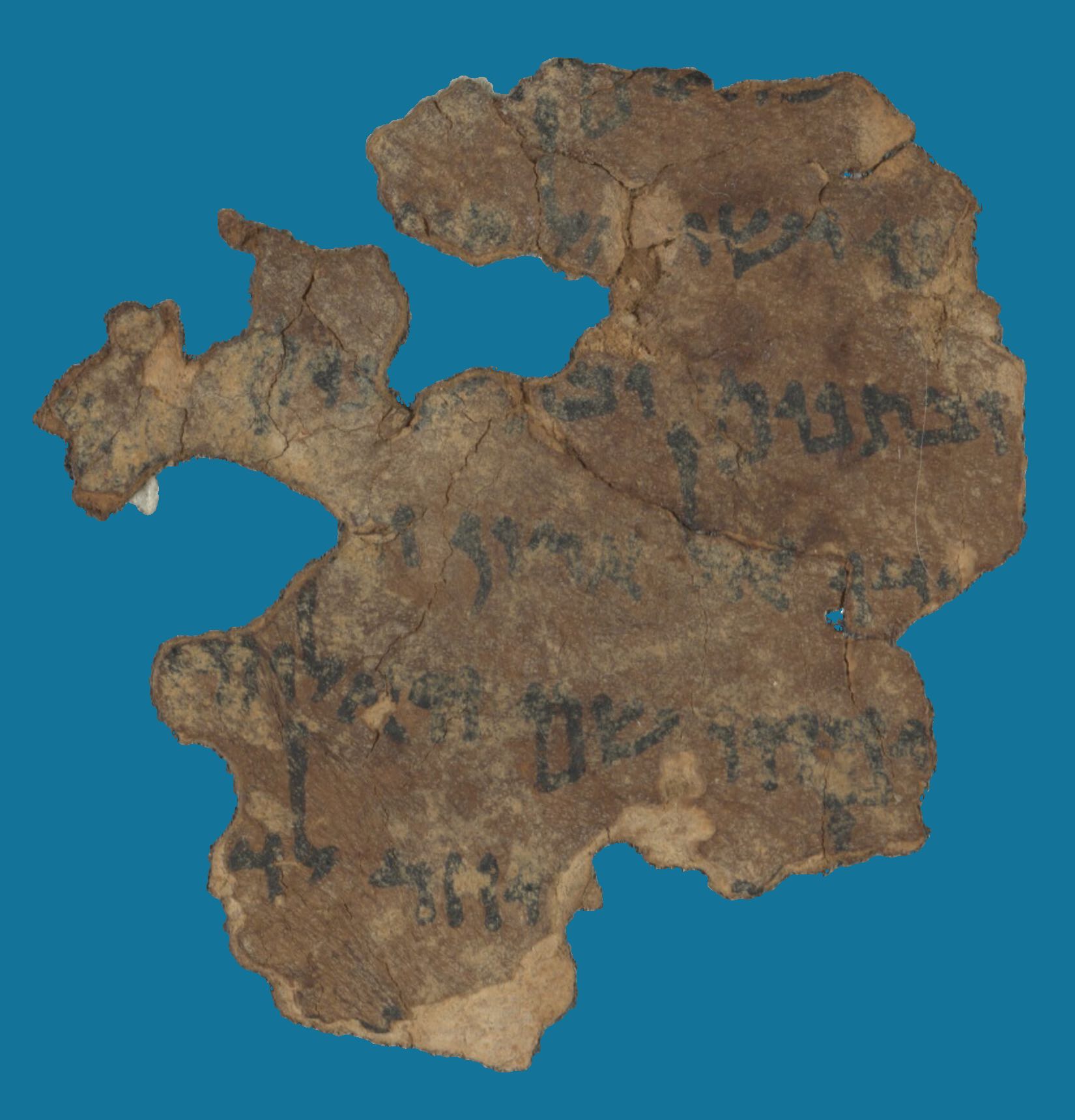}\\
\end{tabular}
\caption{Multiple extracted fragments. Images courtesy of Leon Levy Dead Sea Scrolls Digital Library, Israel Antiquities Authority; photo: Shai Halevi.}
  \label{fig:final_fragments_extraction}
\end{figure}

In the IAA images there may be broken off flecks of parchment or papyrus that are attached to the backing substrate, and thus will successfully be segmented by this pipeline. Such very tiny items, however, are not profitably used within our current research goals and are filtered out of the final collection of results. The filtering is based, like the kernel for morphological closing in step 5b, on the PPI of the IAA images, in this case all polygons with an area less than 1000 are ignored (Figs.~\ref{fig:final_fragment_extraction}, \ref{fig:final_fragments_extraction}).

%\todo{Show the  results step by step on the same images}
\subsection{Evaluation}
\begin{figure}[t]
  \centering
\includegraphics[draft=false,width=0.95\linewidth]{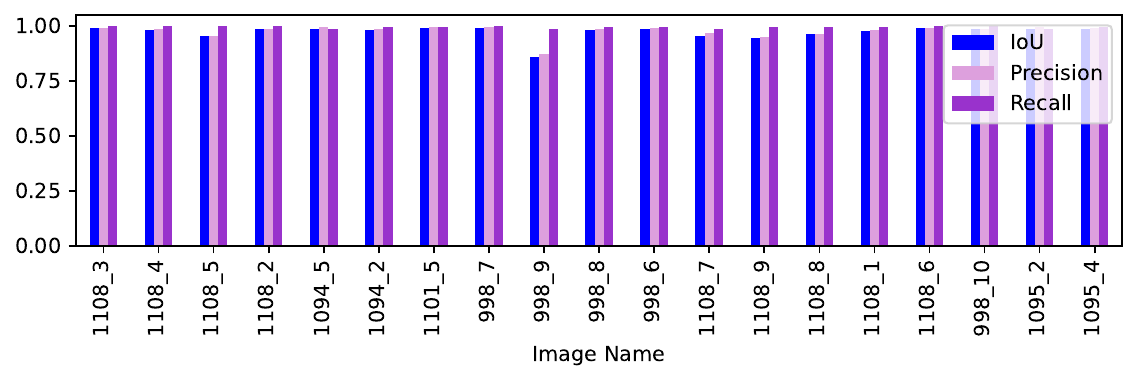}
  \caption{Each bar represents the performance of a single fragment image segmentation, with different colors representing different metrics: IoU, precision, recall.}
  \label{fig:individual_evaluation}
\end{figure}

Fragment segmentation can be considered a semantic segmentation problem because the goal is to segment the fragments from the background. Each pixel of the image is assigned to a semantic class, fragment or background. Therefore we use evaluation metrics for semantic segmentation to evaluate our fragment segmentation results. We measure each metric (IoU, precision, recall, F1, and accuracy) for each fragment image individually (Fig.~\ref{fig:individual_evaluation}) and then calculate the average of these metric values to get the overall performance (Table~\ref{tab:average_evaluation}) of the fragment segmentation. This helps us understand how well the results are on the entire dataset and identify any potential issues with the individual results.

The mean Intersection over Union (IoU) was $0.97$, which indicates a high overlap between the predicted and ground truth binary masks.
The mean precision was $0.98$ and the mean recall was $0.99$, meaning that the model was able to detect most of the fragment pixels present in the ground truth.
The reason that the recall is lower than precision is the tendency to include some background pixels as fragment pixels.
For example, in the case of fragment 998\textunderscore9, the relatively lower precision and recall (Fig.~\ref{fig:individual_evaluation}) values can be attributed to the presence of some background pixels (with Japanese paper) being returned as fragment pixels.

\begin{table}[t]
\caption{The mean values of IoU, precision, recall, F1, and accuracy, calculated for each individual fragment image.}
\label{tab:average_evaluation}
\centering
\begin{tabular}{p{2cm}p{2cm}p{2cm}p{2cm}p{2cm}}
\hline
IoU    & Precision & Recall & F1     & Accuracy \\ \hline
$0.9722$ & $0.9773$    & $0.9947$ & $0.9857$ & $0.9993$   \\ \hline
\end{tabular}
\end{table}

\section{Summary}

The nature of the Israel Antiquities Authority (IAA) digitization of the Dead Sea Scrolls in their collection presents several difficulties for standard approaches to automatically segment the manuscript fragments pictured from the background and metadata elements contained in the image. We propose that a pipeline can be used in such cases in order to isolate each difficulty and solve it using the best, custom-tailored approach. The results of each step are aggregated in specific ways as to reach the desired output. The individual analytical steps outlined here may be useful for other collections of manuscript images that would benefit from segmentation. Further, the usage of a multi-step pipeline will surely be helpful from a conceptual standpoint for other image segmentation projects that encounter problems that have proven intractable when applying any of the more commonly used segmentation techniques. Finally, we employed a ground-truth corpus in order to provide a qualitative evaluation of our automated segmentation results.

\section{Further Research}

\begin{figure}[t]
  \centering
\includegraphics[draft=false,width=0.4\textwidth]{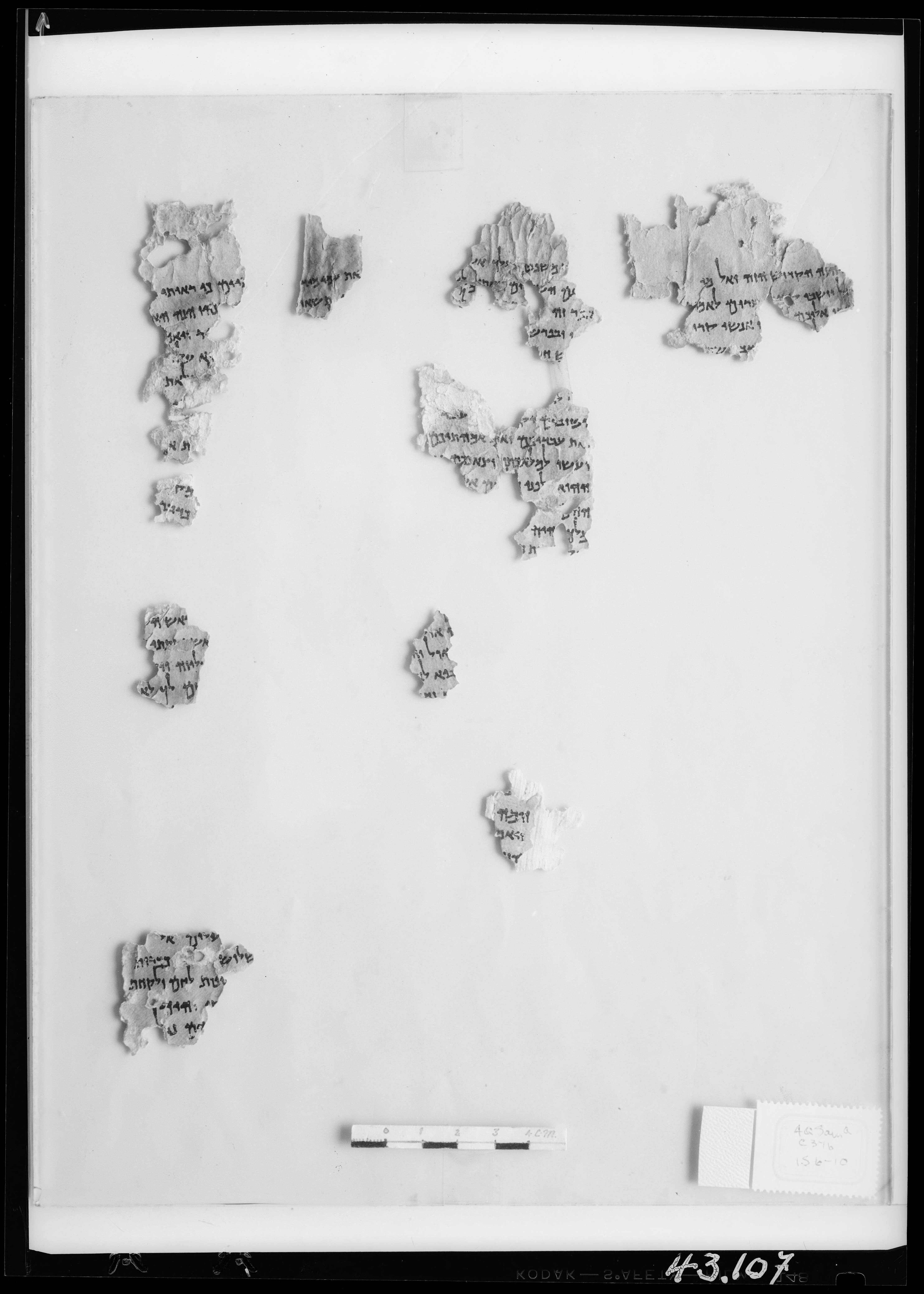}
\caption{PAM series image. Images courtesy of Leon Levy Dead Sea Scrolls Digital Library, Israel Antiquities Authority; photo: Najib Anton Albina.}
  \label{fig:pam_example}
\end{figure}

Considering the success of our experiments on the evaluation dataset and some expected improvements,
the plan is for the pipeline described herein to be applied in the near future to the whole of the IAA corpus of color images and for the results to be incorporated within its online platform.
The resulting images can then be used for some of the following purposes:
\begin{enumerate*}[label=(\alph*)]
\item to rectify the images so that they are upright
\item to accurately place fragments on a canvas in reconstruction attempts
\item to search for multiple images of the same fragment in other sets of images---such as the much older Palestine Archaeological Museum (PAM) images~\cite{PAM} (Fig.~\ref{fig:pam_example}) (cf.\@ \cite{8354133,TK})
\item to allow for the
registration of older images with the new ones (cf.\@ \cite{8354133,TK});
or
\item to align transcriptions with fragments at the glyph-level (cf.\@ \cite{Align}).
\end{enumerate*}

Segmenting fragments on the PAM images would present additional problems, as they are available only as black and white infrared images of plates that usually contain multiple fragments, often with shadows.
See~\cite{8354133} for a first attempt.

\section*{Acknowledgments}
The dataset used in this study is licensed under the Creative Commons Attribution-NonCommercial 4.0 International (CC-BY-NC 4.0) license. All of our derivative images must also be cited according to the paper citation.

\FloatBarrier
\bibliographystyle{plainurl}
\bibliography{SQE}
\end{document}